\documentclass[10pt,journal,compsoc]{IEEEtran}
\ifCLASSOPTIONcompsoc
  \usepackage[nocompress]{cite}
\else
  \usepackage{cite}
\fi
\usepackage{amsmath,amsfonts}
\usepackage{amssymb}
\usepackage{algorithmic}
\usepackage{algorithm}
\usepackage{array}
\usepackage{textcomp}
\usepackage{stfloats}
\usepackage{url}
\usepackage{verbatim}
\usepackage{multirow}
\usepackage{cite}
\usepackage{float}
\usepackage{enumitem}

\usepackage{mathtools}

\def\mathbi#1{\textbf{\em #1}}
\usepackage{subfig}

\DeclareMathOperator*{\argmax}{argmax}

\ifCLASSINFOpdf
   \usepackage[pdftex]{graphicx}
   \DeclareGraphicsExtensions{.pdf,.jpeg,.png}
\else
\fi
\begin{document}
%
\title{Adversarial Attacks Assessment of Salient Object Detection via Symbolic Learning}
%
%
%
%

\author{Gustavo~Olague,~\IEEEmembership{Senior Member,~IEEE,}
		Roberto Pineda,
		Gerardo Ibarra-Vazquez,
		Matthieu Olague,
		Axel Martinez,        
        Sambit~Bakshi,~\IEEEmembership{Senior Member,~IEEE,}
		Jonathan Vargas,        
        and~Isnardo~Reducindo
\IEEEcompsocitemizethanks{\IEEEcompsocthanksitem G. Olague is with the Department of Computer Science, CICESE Research Center, Carretera Ensenada-Tijuana 3918, 22860, M\'{e}xico.\protect\\
Email: olague@cicese.mx
\IEEEcompsocthanksitem R. Pineda is with the Department of Computer Science, CICESE Research Center, Carretera Ensenada-Tijuana 3918, 22860, M\'{e}xico.
\IEEEcompsocthanksitem G. Ibarra-Vazquez is with the Institute for Future of Education. Monterrey Institute of Technology and Higher Education, Monterrey, N.L., 64849, M\'{e}xico.
\IEEEcompsocthanksitem M. Olague is with An\'{a}huac University, Quer\'{e}taro, El M\'{a}rquez, 72246, M\'{e}xico. email: matthieu.olague03@anahuac.mx
\IEEEcompsocthanksitem A. Mart\'{\i}nez is with the Department of Computer Science, CICESE Research Center, Carretera Ensenada-Tijuana 3918, 22860, M\'{e}xico.
\IEEEcompsocthanksitem S. Bakshi is with Department of Computer Science and Engineering, National Institute of Technology Rourkela, Odisha 769008, India. email: bakshisambit@ieee.org
\IEEEcompsocthanksitem J. Vargas is with Terra-Peninsular and Cornell Lab of Ornithology, M\'{e}xico.
\IEEEcompsocthanksitem I. Reducindo is with the Information Science Faculty of Universidad Aut\'{o}noma de San Luis Potos\'{\i}, M\'{e}xico.}
\thanks{Manuscript received April 19, 2005; revised August 26, 2015.}}

%
%

\markboth{Journal of \LaTeX\ Class Files,~Vol.~14, No.~8, August~2015}%
{Olague \MakeLowercase{\textit{et al.}}: Adversarial Attacks Assessment of Salient Object Detection via Symbolic Learning}
%



\IEEEtitleabstractindextext{%
\begin{abstract}
Machine learning is at the center of mainstream technology and outperforms classical approaches to handcrafted feature design. Aside from its learning process for artificial feature extraction, it has an end-to-end paradigm from input to output, reaching outstandingly accurate results. However, security concerns about its robustness to malicious and imperceptible perturbations have drawn attention since its prediction can be changed entirely. Salient object detection is a research area where deep convolutional neural networks have proven effective but whose trustworthiness represents a significant issue requiring analysis and solutions to hackers' attacks. Brain programming is a kind of symbolic learning in the vein of good old-fashioned artificial intelligence. This work provides evidence that symbolic learning robustness is crucial in designing reliable visual attention systems since it can withstand even the most intense perturbations. We test this evolutionary computation methodology against several adversarial attacks and noise perturbations using standard databases and a real-world problem of a shorebird called the Snowy Plover portraying a visual attention task. We compare our methodology with five different deep learning approaches, proving that they do not match the symbolic paradigm regarding robustness. All neural networks suffer significant performance losses, while brain programming stands its ground and remains unaffected. Also, by studying the Snowy Plover, we remark on the importance of security in surveillance activities regarding wildlife protection and conservation.
\end{abstract}

\begin{IEEEkeywords}
Adversarial examples, Visual attention, Brain programming, Adversarial patch, Wildlife conservation.
\end{IEEEkeywords}}

\maketitle

\IEEEdisplaynontitleabstractindextext

%
\IEEEpeerreviewmaketitle

\IEEEraisesectionheading{\section{Introduction}\label{sec:introduction}}
\IEEEPARstart{A}{dversarial} robustness is a critical feature that recent deep learning (DL) architectures are searching for due to their inability to defend themselves from adversarial attacks (AAs), which are strategies that attempt to fool them. Therefore, despite the enormous advantages DL has brought to several research areas, including visual attention (VA), they are untrustworthy. Adversarial examples (AEs) are maliciously-constructed inputs that fool machine learning models by usually degrading the output performance of image classification.

Robustness is a required property when security, safety, and certainty are mandatory in surveillance, life conservation, and defense applications. Robustness is part of a new wave of computer science that is not alien to evolutionary computation \cite{Pineda-BProbustness-2022}. Robustness captures whether the system can deliver correct service conditions beyond the typical domain of operation and without fundamental changes to the original system, i.e., the ability to resist shocks without adapting or changing its behavior.

\subsection{Related Work}

Recent work has shown that these attacks are generalized to salient object detection (SOD) \cite{Li-ROSA-2020}. Li et al. have made the problem of AAs evident since VA is usually an input for more complex visual tasks. The performance of highlighting conspicuous objects is seriously affected by corrupt input images (AEs) in such a way that succeeding stages produce poor results, compromising the entire system. This first work describing the lack of robustness shows how DL compromises the application of such technologies in situations where security and reliability are critical features to avoid failure when detecting the object of interest. Nevertheless, this first work considers only non-target attacks leaving aside other methodologies regarding AAs \cite{Che-2021}. In Section \ref{Problem}, we review a scheme to classify AAs and introduce later the experimental design strategy that we conduct in this work.

In the most recent survey of SOD, the authors included a small section about the lack of robustness against AEs \cite{wenguan_sod_2021}. Wang et al. provided some experiments showing that VA based on deep SOD models is brittle since their performance degrades when contaminated with corrupted inputs. In the first set of experiments, the authors contrasted the behavior of three deep models against heuristic methods. According to their results, heuristic methods do not match deep models in terms of performance; hence, comparing both approaches from a robustness standpoint against AEs is pointless. 

Among the three deep models, the authors highlight the excellent robustness of Pixel-wise Contextual Attention for Saliency Detection (PICANet) against a wide range of input perturbations, including Gaussian blur, Gaussian noise, and rotation \cite{liu_picanet_2018}. Authors attribute this result to its effective non-local operation, revealing that effective network designs improve robustness to random perturbations. Later, in the experimental section, we challenge this idea by showing that this network exhibits poor robustness to Gaussian, Salt \& Pepper, and Speckle noise. In \cite{wenguan_sod_2021}, authors presented a second set of experiments by adopting and modifying an AA algorithm for semantic segmentation to evaluate robustness. They kept the same three deep SOD models for the analysis while excluding the heuristic methods. The results confirm the poor robustness against AEs since attacks designed with the knowledge of deep SOD models fail to highlight the complete salient objects while wrongly recognizing salient objects as part of the background areas. Such attacks prevent SOD models from producing reliable salient object candidates for other safety-critical applications. However, the proposed attack presents weak transferability between the three different network structures. Again, our results contradict this idea, as reported in the experimental section, by showing that many designed attacks affect networks with different architectures.

As we have observed from the reviewed literature, AAs in SOD are rarely studied. Nevertheless, this topic has the potential to become mainstream in computer science as one of the current studies on the susceptibility of deep neural networks for classification tasks \cite{Akhtar2018}. In fact, in this study, we incorporated the most recent work of DL for SOD considering AAs \cite{Wang2023}, which we introduce in Section \ref{deep-SOD}. Currently, SOD plays a significant role in several security applications for detecting candidate interest targets, and the fact that a technique from evolutionary computation is reliable and secure against inconspicuous attacks crafted to affect the functioning of a vast range of government and commercial applications is, without a doubt, a research area to be explored in the future. The present study represents the first attempt to show the effectiveness of artificial evolution as a symbolic learning paradigm against malicious threats in the case of SOD.

\subsection{Problem Statement}
\label{Problem}
AA is a relevant topic in the current study of DL models, which it started to attract attention in image classification models where the phenomenon was first discovered. Soon, the subject spread to diverse areas where it has been demonstrated to appear as a vulnerability in DL modeling. In this section, we summarize the explanation of how AAs affect DL models while focusing on VA modeling. Therefore, given a set of images $\mathcal{I} = \{\mathbf{I}_1, \mathbf{I}_2, \ldots \mathbf{I}_n \}$ and their corresponding proto-objects $\mathcal{P} = \{\boldsymbol{\mathcal{P}}_1, \boldsymbol{\mathcal{P}}_2, \ldots \boldsymbol{\mathcal{P}}_n \}$, the deep SOD model establishes a relationship between an image ($\mathbf{I}_i$) and the proto-object ($\boldsymbol{\mathcal{P}}_i$) through the following equation:

\begin{equation}
    \boldsymbol{\mathcal{P}}=f(\mathbf{I})=A(\mathbf{W}^\intercal \mathbf{I}) \;\;\;,
\end{equation}

\noindent
where function $f(\cdot)$ is the deep SOD model, $\mathbf{W}$ are the associated weight parameters, and $A(\cdot)$ is an activation function. 

AAs denote the erroneous behavior of such models when the input image suffers a small change in its pixels by adding a perturbation $\boldsymbol{\rho}$ to create the AE $\mathbf{I}_{\boldsymbol{\rho}} = \mathbf{I} + \boldsymbol{\rho}$ such that:

\begin{equation} \label{eq:aa}
\begin{split}
    f(\mathbf{I}) \neq f(\mathbf{I}_{\boldsymbol{\rho}}) \;\; \mathbf{s.t.}\;\; ||\mathbf{I} - \mathbf{I}_{\boldsymbol{\rho}}||_{p}<\alpha 
\end{split}
\end{equation}

where $p\in N \;|\; p\geq 1$ and $\alpha \in R \;|\; \alpha\geq 0$. The AE is considered the intentionally modified input image that is recognized differently than the $\mathbf{I}$. The level of change in the pixels is defined by $||\mathbf{I} - \mathbf{I}_{\boldsymbol{\rho}}||_{p}<\alpha$ that constrains the AE to be too small so it may be imperceptible to the human eye, although this constraint can be overlooked. A simple explanation of how AAs break DL models, rendering them vulnerable, is that if we compute the dot product with the associated weight matrix from the model. For the AE, we obtained $\mathbf{W}^\intercal \mathbf{I}_{\boldsymbol{\rho}}= \mathbf{W}^\intercal\mathbf{I} + \mathbf{W}^\intercal\boldsymbol{\rho}$. This is caused by the linearity of neural networks, which are intentionally designed with activation functions that perform linearly so that they are easier to optimize. Therefore, the AE will grow the activation function by $\mathbf{W}^\intercal\boldsymbol{\rho}$.

Additionally, the linearity of DL models reveals another consequence of AAs. Every AE that is easy or difficult to compute in a specific model affects another model, even with different architectures. They only have to be trained for the same task. However, we extend this effect between different tasks, as explained in Section \ref{ap}. This phenomenon is defined as the transferability effect that establishes the spread of this erroneous behavior between DL models with such small perturbations. AAs are usually classified depending on their strategy, which defines how the perturbation is created. White box attacks assume the complete knowledge of the architecture model, its parameter values, and the trained task. On the contrary, a black box attack optimizes the perturbation during a testing procedure to change the original prediction without knowing the model. These strategies are also considered to fool a model into believing a specific label (targeted attack) or that the predicted label is irrelevant (untargeted attack) while it is not the original label.

\subsection{Contributions}

This research is part of discovering new properties that evolutionary algorithms possess against DL. We extend the first results reported at the 25th International Conference on the Applications of Evolutionary Computation, where we reported the first discoveries about the robustness against AAs of brain programming (BP) for the problem of SOD \cite{Pineda-BProbustness-2022}. We note the following contributions:



\begin{enumerate}
    \item This study shows the robustness of a symbolic SOD algorithm against AAs by conducting a deeper analysis and contrasting it with five DL models, three AAs, three different noises, and five different datasets. 
    \item This study considers not only standard databases, but a real-world problem focused on wildlife conservation that proves a real challenge for deep-learning SOD methods.
\end{enumerate}


\section{Methodology}

This section contains four subsections. First, we describe six learning-based methods commonly applied to SOD problems. Then, we provide a mathematical formulation for learning-based SOD modeling. Later, we introduce the idea of robustness through continuity criteria. Finally, we list our choice of AAs.

\subsection{Symbolic and Subsymbolic Approaches for SOD}
\label{deep-SOD}

Nowadays, SOD researchers predominantly prefer solutions based on DL in contrast to traditional solutions that lack the idea of learning, so the latter's influence has been reduced recently. However, current research continues the tradition by using the same databases and assessment schemes outlined before the flourishing of DL in VA. BP is a design scheme where we incorporate symbolic learning to enhance traditional methodologies. All research follows the path of empirical investigation while modeling hand-designed solutions regardless of the approach. In conventional methods, features were designed by hand, and in DL methods, architectures are still designed by hand. Here, we propose to compare five different networks with a recently designed algorithm under the BP scheme, the results of which we will describe later. The five selected networks have similar characteristics since all architectures are fully convolutional networks, adopt a fully supervised learning, bottom-up or top-down network scheme, and follow single-task learning.

\begin{figure*}[!ht]
\centering
\includegraphics[width=5.7in]{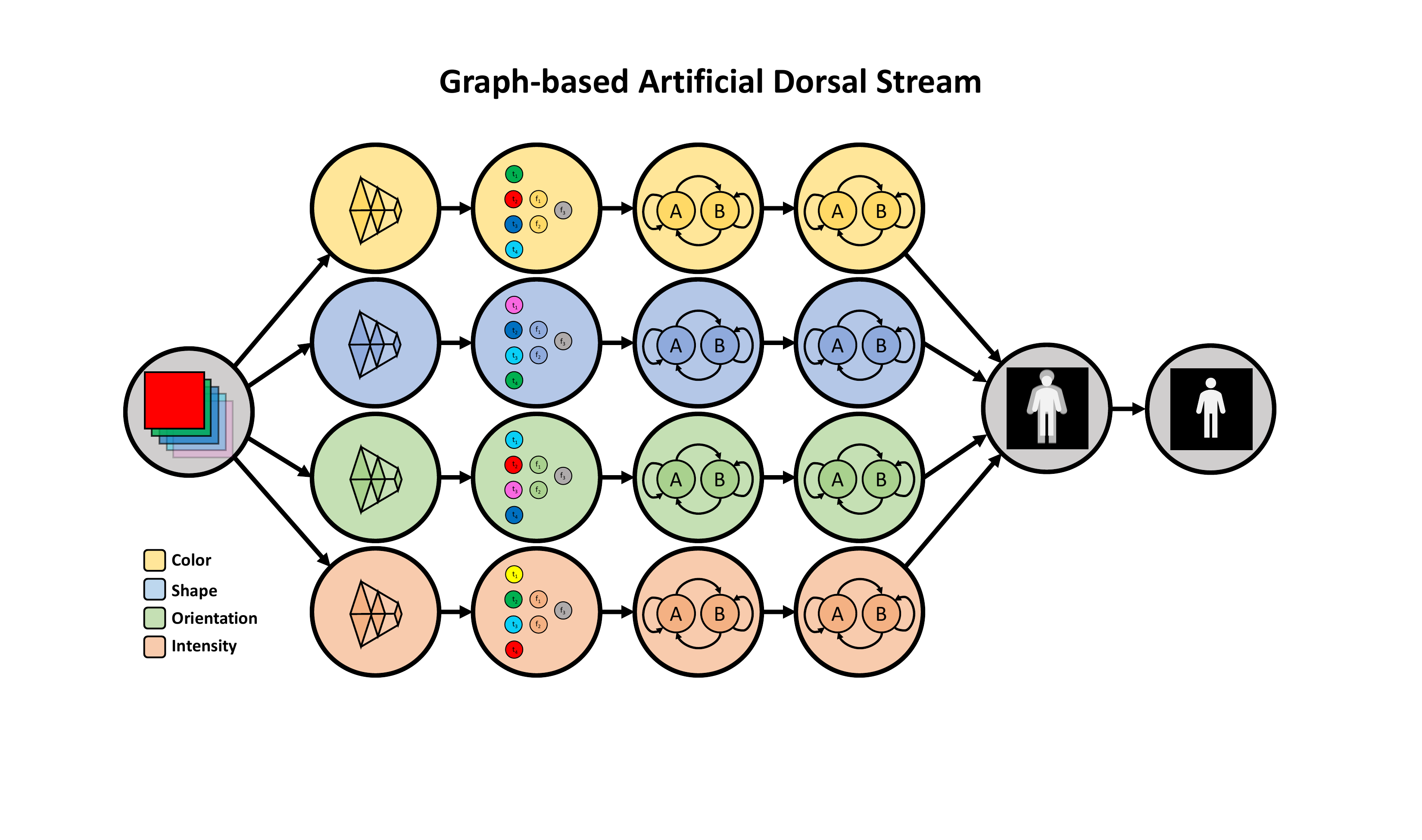}%
\caption{The proposed method encodes an individual as a template, and we apply artificial evolution to discover a set of mathematical and computational functions optimized for the task of salient object detection. The evolved solutions are robust against different AEs, as the experiments exemplify.}
\label{fig_sim}
\end{figure*}

\begin{itemize}
    \item \emph{BP} is an evolutionary computation paradigm loosely based on the inner workings of the visual cortex. In this paper, we focus on the analogy to the dorsal stream to create computational models of SOD from an input image (Figure \ref{fig_sim}). The goal is to show the robustness of the methodology against AEs. The strategy follows a goal-oriented framework where we study learning as a symbolic optimization process where an individual consists of a template describing the VA model while discovering critical parts of the algorithm through artificial evolution. This study focuses on the mathematical notation to formulate the SOD problem from the viewpoint of robustness and leaves the full explanation of the algorithm to consult in a previous document \cite{Olague-SOD-2022}.
    \item Adversarial Robust SOD Networks with Learnable Noise (LeNo) proposes the usage of simple shallow noise and noise estimation embedded in the encoder and decoder of arbitrary SOD networks to defend against AAs \cite{Wang2023}. Introduced in 2022, LeNo initializes the shallow noise with a cross-shaped Gaussian distribution inspired by the center prior of the human VA mechanism. Instead of adding additional network components for post-processing, the proposed noise estimation modifies only one decoder channel. This method outperforms previous works on adversarial and clean images, contributing to the stronger robustness of SOD.
    \item SOD via Extremely-Downsampled Network (EDN) focuses on enhancing high-level features through extreme downsampling to effectively learn a global view of the whole image, leading to accurate salient object localization \cite{Wu2022}. Proposed in 2022, improved multi-level feature fusion is accomplished through the construction of the scale-correlated pyramid convolution to build a decoder for recovering object details from the above extreme downsampling. Not only does this work achieve state-of-the-art performance, but it also focuses on working at real-time speeds.
    \item Boundary-Aware SOD (BASNet) is a deep convolutional neural network (CNN) focusing on the boundary quality that appears in 2019 \cite{qin_basnet_2021}. The proposed BASNet consists of two modules, the first inspired by U-Net and a semantic segmentation model (SegNet), which implement a salient object prediction module as an Encoder Decoder network, capturing high-level global contexts and low-level details at the same time. The input convolutional layer and the first four stages are adopted from \cite{He2015}, the transfer Residual Neural Network (ResNet-34), while the residual refine module consists of filters followed by a batch normalization and a Rectified Linear Unit (ReLU) activation function \cite{agarap2018deep}.
    \item Learning PICANet aims to generate an attention map at each pixel over its context region and construct an attended contextual feature to enhance the feature representability of convolutional networks \cite{liu_picanet_2018}. The authors proposed the network in 2018 based on two pixel-wise attention modes: global attention and local attention. For each location, the former generates attention based on the Recurrent Neural Network over the whole feature map \cite{visin2015renet}, while the latter works on a local region using several convolutional layers. The whole network is based on the VGG 16 layer \cite{Liu2015} as the backbone while applying U-Net, which is a CNN developed for biomedical image segmentation.
    \item The Deep Hierarchical Saliency Network (DHSNet) is an end-to-end CNN detecting salient objects that appeared in 2016 \cite{liu_dhsnet_2016}. It consists of a global view step followed by a Hierarchical Recurrent CNN. Based on the Very Deep Convolutional Networks (VGG net), the first network creates a coarse global saliency map to roughly detect and localize salient objects. The second step refines saliency maps in detail by incorporating local contexts.
\end{itemize}

We have appreciated all five networks based their architectures on previous development of CNNs. The idea of incorporating symbolic learning based on previous neuroscientific research is equivalent; therefore, we aim to show the robustness of symbolic modeling.

\subsection{SOD Evaluation}

The image-based SOD problem can be formulated as follows. Given an input image $\mathbf{I} \in \mathbb{R}^{u\times v\times 3}$ of size $u \times v$ for each color channel, an SOD model $f$ maps the input image $\mathbf{I}$ to a proto-object $\boldsymbol{\mathcal{P}} = f(\mathbf{I}, F, T, a) | \boldsymbol{\mathcal{P}} \in \{0, 1\}^{u \times v}$, where $F$ and $T$ represent the function and terminal sets, respectively, from the feature extraction, and $a$ is the set of parameters controlling the evolutionary process.

BP is the algorithm in charge of tuning $(F, T)$, looking for optimal feature extraction from the input images using the visual operators embedded into the artificial dorsal stream. For learning-based SOD, the meta-model $f(\cdot)$ is learned through a set of training samples. Given a set of images $\mathcal{I} = \{\mathbf{I}_1, \mathbf{I}_2, \ldots \mathbf{I}_n \}$, with $n$ defining the number of training images, and corresponding binary SOD ground-truth masks

\begin{displaymath}
\mathcal{G} = \{ \mathbi{G}_1, \ldots, \mathbi{G}_n | \mathbi{G} \in \{0, 1\}^{u \times v} \},
\end{displaymath}

\noindent
the goal of learning is to find $f \in \mathcal{F}$ that minimizes the prediction error, i.e., $\displaystyle\Sigma_n Q(\boldsymbol{\mathcal{P}}_n, \mathbi{G}_n)$, where $Q$ as a quality criterion in this case is a certain distance measure (e.g., $F_{\beta}$ measure), and $\mathcal{F}$ is the set of potential mapping functions. Deep SOD methods typically model $f(\cdot)$ through modern DL techniques, as we reviewed in Section \ref{deep-SOD}.

In the case of BP, we have $f(\mathbf{I}, F, T, a)$, such that

\begin{displaymath}
\boldsymbol{\mathcal{P}}^* \in \mathcal{S} : \mathcal{Z}(\boldsymbol{\mathcal{P}}^*, \mathbi{G}) \geq \mathcal{Z}(\boldsymbol{\mathcal{P}}, \mathbi{G})
\end{displaymath}

\noindent
which maximizes the overlap between $\boldsymbol{\mathcal{P}}$ and $\mathbi{G}$ from the solution space $\mathcal{S}$, such that 

\begin{displaymath}
Z = \displaystyle\sum_{i=1} ^{n} \mathcal{Z}(\boldsymbol{\mathcal{P}}_i,\mathbi{G}_i)
\end{displaymath}

\noindent
where $\mathcal{Z}$ is a statistical measure of accuracy, $\mathcal{Z} \triangleq F_{\beta}(\cdot)$ , that is computed as follows:

\begin{equation}
F_{\beta}(P_i, R_i) = \displaystyle\sum_{i=1} ^{n} \displaystyle\sum_{j=1} ^{m} \frac{(1+\beta ^2) p_j \cdot r_j}{\beta^2 p_j + r_j}
\label{F-measure}
\end{equation}

\noindent
with $m$ representing the number of thresholds to build the proto-object, and $n$ defines the number of training images. Precision data of an image are denoted with $P_j = (p_1, p_2,\ldots, p_n)$ and recall data by $R_j = (r_1, r_2,\ldots, r_n)$ with

\begin{displaymath}
\begin{array}{c}
p = \frac{t_p}{t_p + f_p} ,\\
r = \frac{t_p}{t_p + f_n} , 
\end{array}
\end{displaymath}

\noindent
where $t_p$ means true positive, $f_p$ is false positive, and $f_n$ denotes false negative. $F_\beta$ measures the effectiveness of retrieval with respect to a user who attaches $\beta$ times as much importance to recall as precision. $\beta^2$ is set to $0.3$ to emphasize the precision following the standard protocol \cite{li_secrets_2014}.

We empirically calculate precision and recall based on a number $n$ of thresholds for each image in the dataset while considering each algorithm run. The $F_{\beta}$ measure of each of these results is calculated based on the generated saliency map and splicing it with the ground-truth image. After this, we take the maximum value to obtain each image's best $F_{\beta}$ score. Then, we calculate the average of all the images' $F_{\beta}$ scores to obtain the final $F_{\beta}$ value according to Equation (\ref{F-measure}), which corresponds to $Z$.

\subsection{Model Robustness through Continuity Criteria}

Let $\boldsymbol{\mathcal{P}} = f(\mathbf{I})$, where $f(\cdot)$ is the model (NN, BP, etc.). For example, $\boldsymbol{\mathcal{P}} = f(\mathbf{I}, F, T, a) \sim f(\mathbf{I}) \therefore f(\mathbf{I}) \neq f(\mathbf{I}_{\boldsymbol{\rho}}) : ||\mathbf{I} - \mathbf{I}_{\boldsymbol{\rho}}|| < \alpha$.

AEs find input $\mathbf{I}_{\boldsymbol{\rho}}$ in the subspace $\mathcal{I}'$, such that $\mathbf{I}_{\boldsymbol{\rho}} \in \mathcal{I}'$ and $f(\mathbf{I}) \neq f(\mathbf{I}_{\boldsymbol{\rho}})$. Nevertheless we denote robustness in terms of function continuity. Given a model's function $f()$ in an input subspace. $\mathcal{I}$ is said to be robust at $\mathbf{I} \in \mathcal{I}$; if $\mathbf{I}_{\boldsymbol{k}} \mapsto \mathbf{I}$, then $f(\mathbf{I}_{\boldsymbol{k}}) \mapsto f(\mathbf{I})$. Equivalently, $f(\mathbf{I})$ is robust at $\mathbf{I}$, for all $\mathbf{I}' \in \mathcal{I}$, if given a $\epsilon > 0$, there is a $\delta > 0$, such that $||\mathbf{I} - \mathbf{I}'|| < \delta$ implies $||f(\mathbf{I}) - f(\mathbf{I}')|| < \epsilon$. Hence, if $f(\mathbf{I})$ is robust for every $\mathbf{I}$, then $f(\mathbf{I})$ is said to be robust on $\mathcal{I}$.

\subsection{AAs}
\subsubsection{Fast Gradient Sign Method (FGSM)}

FGSM was first implemented by Goodfellow et al. by noting that linear behaviors of high-dimensional spaces easily generate AEs \cite{goodfellow_2015}. FGSM proposes to increase the loss of the detector by solving the following equation: $\boldsymbol{\rho} = \epsilon \text{ sign}(\Delta J(\theta, \mathbf{I}, \mathbi{G}))$, where $\Delta J()$ computes the gradient of the cost function around the current value of the model parameters $\theta$ with respect to the image $\mathbf{I}$, and the ground truth $\mathbi{G}$. $\text{sign}()$ denotes the sign function, which maximizes the magnitude of the loss and $\epsilon$ is a small scalar value that restricts the norm $L_{\infty}$ of the perturbation.

\subsubsection{Multipixel Attack}
The multipixel adversarial perturbation is a black box untargetted attack since it does not require network information. The one-pixel attack is the basis for this algorithm \cite{Su2017}. However, the original algorithm is not suitable for real-world problems since it can only work for icon images. In summary, let the vector $\mathbf{I} = (\mathbf{I}_1,\ldots,\mathbf{I}_n)$ be an n-dimensional image, which is the input of the salient object detector $f()$ that correctly predicts the object $t$ from the image. The statistics of $\mathbf{I}$ associated with the object $t$ is $\mathcal{Z}(f(\mathbf{I}), f(\mathbf{I} + e(\mathbf{I})))$. It builds an additive adversarial perturbation vector $e(\mathbf{I}) = (e_1,\ldots,e_n)$ according to $\mathbf{I}$, the salient object detector, and the limitation of maximum modifications $d$, a small number that expresses the dimensions that are modified, while other dimensions of $e(\mathbf{I})$ that are left are zeros. The main purpose is to find the optimal solution $e(\mathbf{I})^*$ that solves the following equation:

\begin{displaymath}
\begin{array}{c}
\displaystyle\min_{e(\mathbf{I})^*} \mathcal{Z}(f(\mathbf{I}), f(\mathbf{I}) + e(\mathbf{I}))  \\
\text{s.t.} ||e(\mathbf{I})||_0 \leq d
\end{array}
\end{displaymath}

\subsubsection{Adversarial Patch (AP)} \label{ap}

The AP is a white box attack that creates a particular form of perturbation, which builds a universal perturbation with a patch shape \cite{brown_2018}. Even though this attack is performed to be used on image classification tasks, its universality makes it perfect to be analyzed with the transferability effect that this perturbation provokes. The AP builds the perturbation, maximizing $f_{target}(\mathbf{I}+\mathbf{\hat{p}}))$ to a specific class where it finds the optimal patch $\mathbf{\hat{p}}$. Universal perturbations such as the AP pose a powerful foolproof that relies on the wide variety of transformations that can be applied to the patch to fool the system, or it can even be printed to work in real-world conditions.   

The algorithm trains the AP $\mathbf{\hat{p}}$ using a set of images $\mathcal{I}$, and a variant of the expectation over transformation framework to optimize the following equation:

\begin{equation}
    \mathbf{\hat{p}}= \argmax_{\mathbf{\hat{p}}} \mathbb{E}_{\mathbf{I}\in \mathcal{I}. t \in T. l \in L}[\log f(y_{target},A(\mathbf{p},\mathbf{I},l,t))] \;\;\;,
\end{equation}
\noindent

where $y_{target}$ represents the target class in the image classification model and $A(\mathbf{p}, \mathbf{I}, l, t)$ is a function that first applies the transformation $t$ from a distribution over transformations $T$ to the patch $\mathbf{p}$. Next, it puts the transformed patch $\mathbf{p}$ at the location $l$ from a distribution over locations $L$ to the image $\mathbf{I}$. The effectiveness of these patches resides in the expectation over the training images that increases success, regardless of the background.

\subsubsection{Noise-based Attacks}

AEs are usually interpreted as strategically perturbed images that fool DL models. However, randomly perturbed images are also studied to verify the robustness of SOD models. We employ three types of noise (Gaussian, Salt \& Pepper, and Speckle noise) to study the behavior of such algorithms with randomly perturbed inputs. Gaussian noise adds a white noise with a probability density function of a normally distributed random variable with an expected value $\mu$ and variance $\sigma$. The general probability density function is as follows:

\begin{equation}\label{eq_gaussian_distribution}
   g(x) = \frac{1}{\sigma\sqrt{2\pi}}e^{\frac{-(x-\mu)^2}{2\sigma^2}}
\end{equation}

\noindent
Salt \& Pepper noise with density ($d$) assigns each pixel ($p$), from the total pixels ($tp$) in an image, a random probability value ($pv$) from a standard uniform distribution between $(0,1)$. Therefore, it applies to the following cases 

\begin{itemize}
    \item $p = 0$, for $pv \in (0,d/2)$ limited to $d \times tp/2$ pixels.
    \item $p = \max$ value, for $pv \in [d/2,d)$ limited to $d \times tp/2$ pixels.
    \item $p=p$, for $pv \in [d,1)$. 
\end{itemize}

\noindent    
Speckle noise add a multiplicative perturbation using the following equation $J = \mathbf{I}+m \times \mathbf{I}$, where $m$ is a uniformly distributed random noise with $\mu=0$ and $\sigma=0.05$.

\section{Experimental Results}
We organize this section into three subsections describing the databases, the generation of AEs, and validation results.

\subsection{Databases}

In this work, we use five different image databases as a way to evaluate adversarial robustness. Four of these databases show common everyday objects. However, we experimented with a new image database not used in previous SOD research that holds real-world objects. By ``\textit{real-world}," we refer to objects in a scene positioned in their natural and unperturbed environment and occurring naturally without any external factors. A good example is nature photography, where photographers capture an animal without manipulating the scene. Data collection normally includes respecting the many properties that might be present when taking a photograph, i.e., light and weather conditions, debris, low contrast, obstructions, and object position, among others. Experimenting with a real-world database is a big step toward SOD, given that many research papers focus on using images specifically designed to test a novel algorithm and where the properties for detection are mostly advantageous \cite{wenguan_sod_2021}. These databases mostly show large objects with a high contrast between the foreground and background. The databases used in this work that refer to the point just mentioned are FT \cite{ahcnata_ft_2009}, ImgSal \cite{li_imgsal_2013}, PASCAL-S \cite{li_secrets_2014}, and DUTS \cite{Wang2017}.

\begin{figure}[!h]
    \centering
    \includegraphics[width=1.0\linewidth]{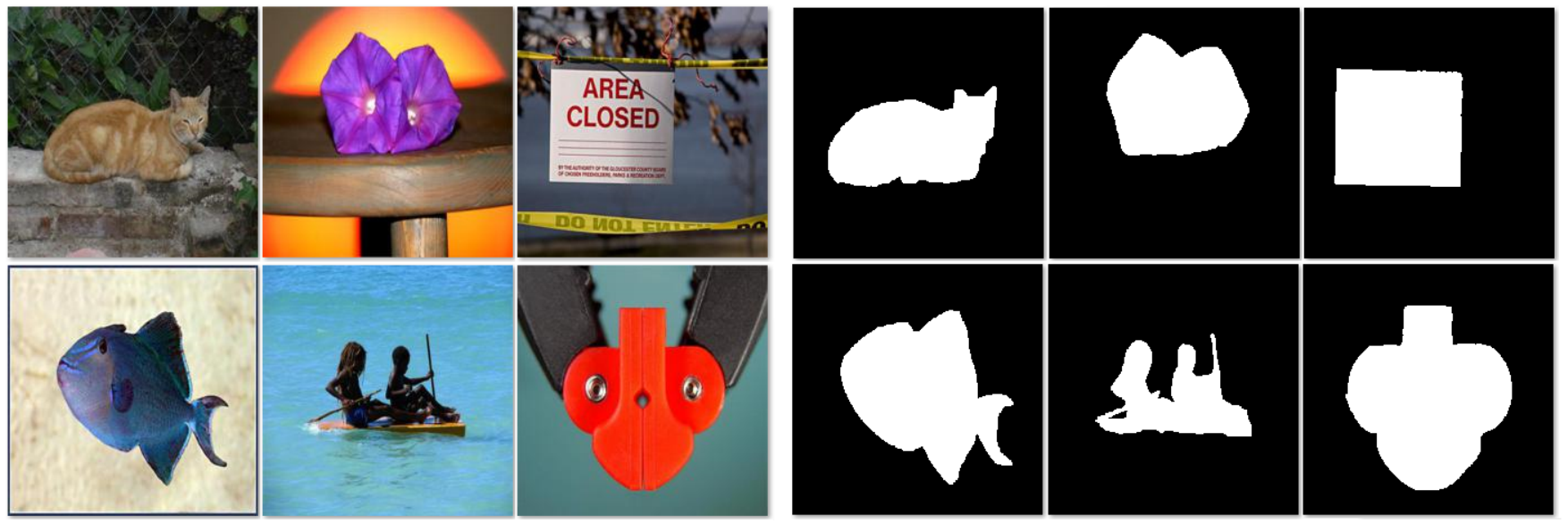}
    \setlength{\belowcaptionskip}{10pt}
    \caption{Example images of the FT database with the corresponding ground-truth.}
    \label{fig:ft_examples}
\end{figure}

\begin{figure}[!h]
    \centering
    \includegraphics[width=1.0\linewidth]{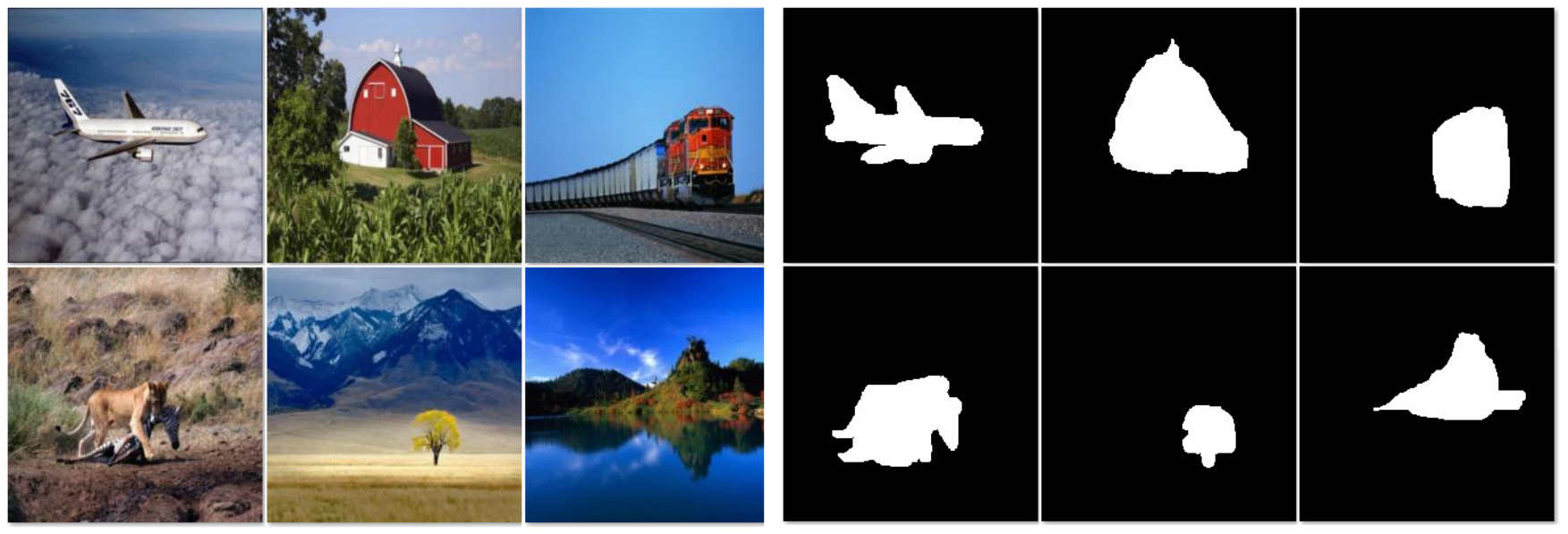}
    \setlength{\belowcaptionskip}{10pt}
    \caption{Example images of the ImgSal database with the corresponding ground-truth.}
    \label{fig:imgsal_examples}
\end{figure}

\begin{figure}[!h]
    \centering
    \includegraphics[width=1.0\linewidth]{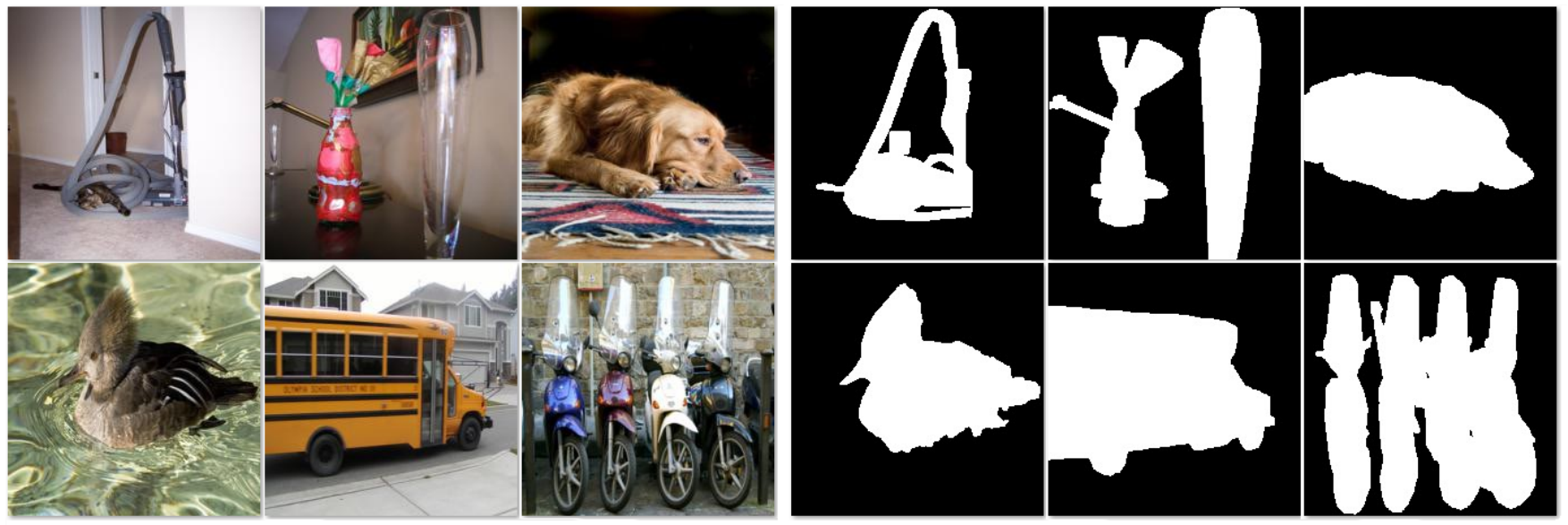}
    \setlength{\belowcaptionskip}{10pt}
    \caption{Example images of the PASCAL-S database with the corresponding ground-truth.}
    \label{fig:pascal_examples}
\end{figure}

\begin{figure}[!h]
    \centering
    \includegraphics[width=1.0\linewidth]{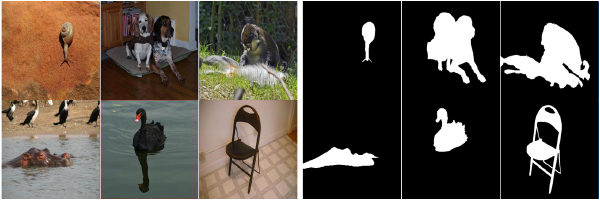}
    \setlength{\belowcaptionskip}{10pt}
    \caption{Example images of the DUTS database with the corresponding ground-truth.}
    \label{fig:duts_examples}
\end{figure}

FT contains 800 images for training and 200 for testing, and Figure \ref{fig:ft_examples} provides some images of the dataset. The dataset includes diverse portraits of animate and inanimate objects with image sizes ranging from $324 \times 216$ up to $400 \times 300$ pixels. PASCAL-S consists of 680 images for training and 170 images for testing (Figure \ref{fig:imgsal_examples}). PASCAL-S contains scenes of domestic animals, persons, and means of air and sea transport, with images ranging from $200 \times 300$ to $375 \times 500$ pixels. IMGSAL provides 235 images, split into 188 training images and 47 images for testing (Figure \ref{fig:pascal_examples}). This dataset includes wild animals, flora, and different objects and people with image sizes of $480 \times 640$ pixels. Finally, regarding standard datasets, DUTS is comprised of 10,553 training images and 5,019 test images (Figure \ref{fig:duts_examples}). DUTS portrays people, animals, insects, and objects in a wide variety of scenarios with image sizes ranging from $266 \times 400$ to $400 \times 192$ pixels.

\begin{figure}[!h]
    \centering
    \includegraphics[width=1.0\linewidth]{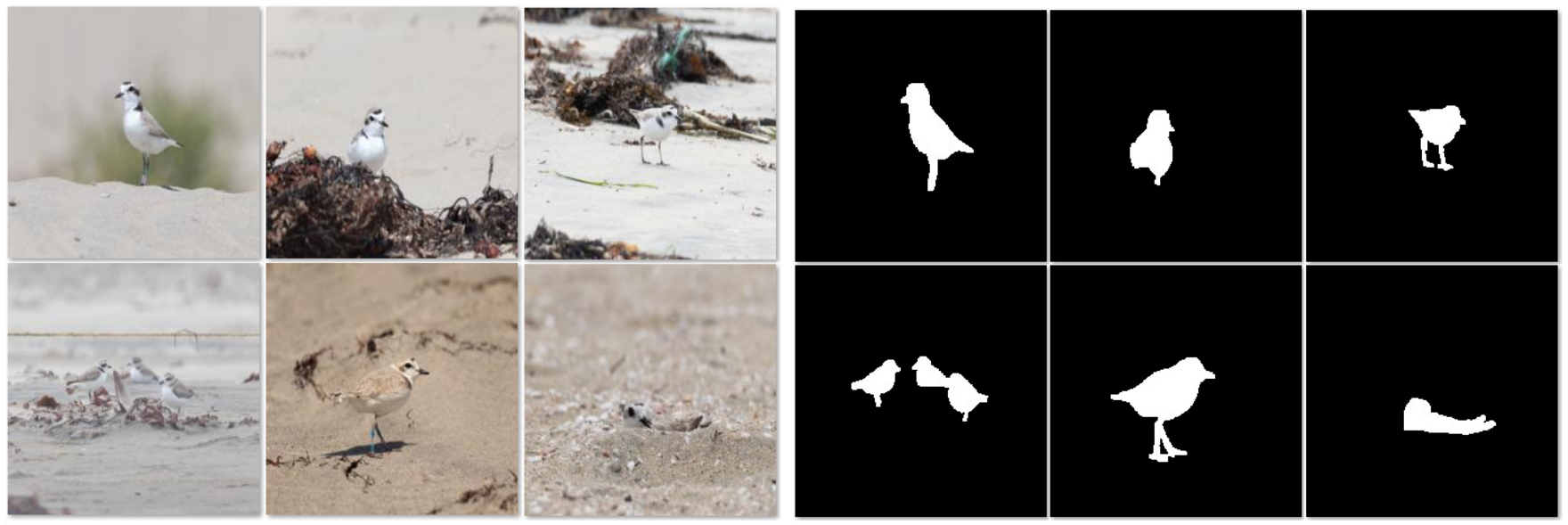}
    \setlength{\belowcaptionskip}{10pt}
    \caption{Example images of the SNPL database with the corresponding ground-truth.}
    \label{fig:snpl_examples}
\end{figure}

This work has another purpose: for experimenting with AAs using a live bird database. The Snowy Plover (Charadrius nivosus) was listed by the U.S. Fish \& Wildlife Service (USFWS) as a threatened species under the federal Endangered Species Act. The Snowy Plover is a small shorebird that can reach a length of 6.7 in (17 cm) with a wingspan of 13.4 in (34 cm). The bird has a short, thin, black bill and gray legs. The upper body varies from grayish to light-brown, with a white belly and black on the forehead and ears. The Snowy Plover inhabits sandy beaches. Many organizations have been working for the protection of the Snowy Plover, incorporating mainstream technology based on DL. However, AAs expose the vulnerabilities of many neural networks used for wildlife conservation. These neural networks are part of real-life monitoring systems such as camera traps for animal identification, and counting \cite{norouzzadeh_2018}, drones for scanning a habitat \cite{doull_2021}, and even acoustic recording devices \cite{kahl_birdnet_2021}. Many of these applications involve defending against the attacks of individuals performing illegal activities that put wildlife in danger, such as poaching, which has become an increasing problem in today's world and a subject that conservationists desperately need a solution to, not only regarding accuracy but reliability and security. Other activities threatening species' well-being are the constant increase in human population and their disturbance of the animal's natural habitat, such as beaches where the Snowy Plover and many other birds reside. By exposing the vulnerabilities of these neural networks, the scientific community can focus on improving the resilience of their algorithms to prevent attacks by malicious individuals. This work shows that an evolutionary computation technique is appropriate to approach SOD, with the advantage of not being susceptible to AAs while performing high detection. 

The fourth real-world database used in this work comprises 250 images of the Snowy Plover, named the SNPL database (Figure \ref{fig:snpl_examples}). We took images using a Nikon DSLR camera and a 200-500mm f/5.6 Supertelefoto lens, considering different weather, exposure, and range conditions. We took each of the 250 photographs in the bird's natural habitat, consisting of sandy beaches and mudflats on the Pacific coast of Baja California, Mexico. Each one of these images shows one or more Snowy Plovers, whose size on the scene varies according to the distance at which we took a photograph while adjusting the focal length. Another essential feature involves the background, where a variety of distractors, such as plants, water bodies, and small objects (shells and debris), are present. Also, many of these images contain occlusions that partially cover the (ground truth) object of interest. Note also that the animal's sandy beach habitat and the primary color of its feathers have similar color tones. Many of these images lack a high contrast between the background and the Snowy Plover. Due to these characteristics, the SNPL database is a great basis to study such a real-world problem. This dataset represents a challenging scenario to test different SOD models. Also, we can compare it with the previous datasets to understand the limitations of academic datasets whose optimal size and exposure conditions prevent us from understanding the effect of AEs on VA.

\subsection{Generation of AEs}

In order to test the robustness, or resistance of the detection algorithms to AAs, including BP, we generated AEs for the validation set of each database used in this work. We used the Python programming language, version 3.8.3 and the machine learning framework PyTorch, version 1.6.0 \cite{NEURIPS2019_9015} and explained the process for generating each AE.

\begin{figure*}
    \centering
        \subfloat{\includegraphics[width=0.14\textwidth]{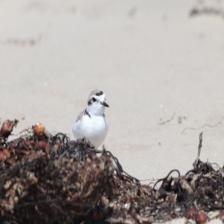} 
        \label{fig:snpl_fgsm0}}
        \hfil
        \subfloat{\includegraphics[width=0.14\textwidth]{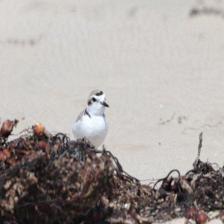}
        \label{fig:snpl_fgsm2}}
        \subfloat{\includegraphics[width=0.14\textwidth]{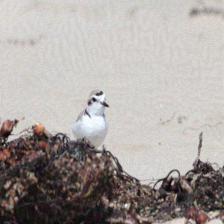}
        \label{fig:snpl_fgsm4}}
        \subfloat{\includegraphics[width=0.14\textwidth]{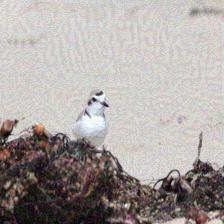}
        \label{fig:snpl_fgsm8}}
        \subfloat{\includegraphics[width=0.14\textwidth]{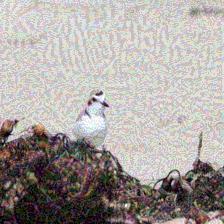}
        \label{fig:snpl_fgsm16}}
        \subfloat{\includegraphics[width=0.14\textwidth]{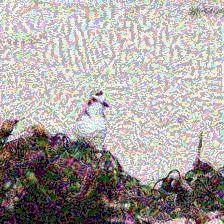}
        \label{fig:snpl_fgsm32}}
        \subfloat{\includegraphics[width=0.14\textwidth]{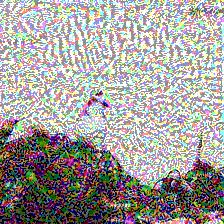}
        \label{fig:snpl_fgsm64}}
    \hfil
    
        \subfloat{\includegraphics[width=0.14\textwidth]{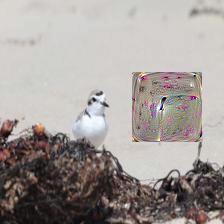} 
        \label{fig:snpl_advpatch_big}}
        \subfloat{\includegraphics[width=0.14\textwidth]{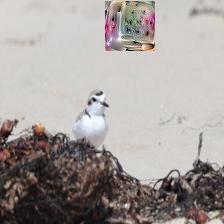}
        \label{fig:snpl_advpatch_small}}
        \subfloat{\includegraphics[width=0.14\textwidth]{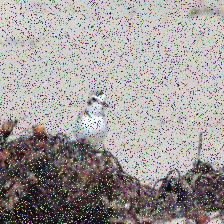}
        \label{fig:snpl_multipixel}}
        \subfloat{\includegraphics[width=0.14\textwidth]{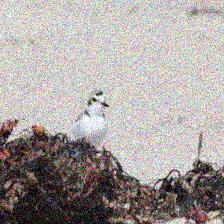}
        \label{fig:snpl_gaussian}}
        \subfloat{\includegraphics[width=0.14\textwidth]{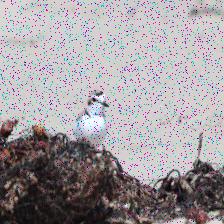}
        \label{fig:snpl_saltpepper}}
        \subfloat{\includegraphics[width=0.14\textwidth]{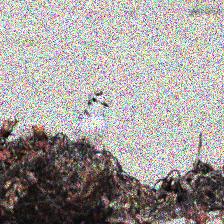}
        \label{fig:snpl_speckle}}
    
    \caption{Perturbations over an image of the SNPL database. On top is the original image, while in the second row we observe the image with different levels of perturbation using FGSM; the third row provides the resulting AEs using two versions of the adversarial patch, the multipixel attack, and the three different noises.}
    \label{fig:snpl_fgsm}
\end{figure*}

We created six sets of AEs for the validation set of each database using FGSM coupled with the training stage of PICANet. We generated each of the six sets with a value of $\epsilon = \{2,4,8,16,32,64\}$, so the images of each set have disturbances with different intensities, with $2$ being the lowest perturbation and $64$ being the highest variation. The second row of Figure \ref{fig:snpl_fgsm} shows examples of this perturbation applied to an image of the Snowy Plover.

Regarding the AP, we created two sets using ResNet-50. The first set contains images where the patch covers an area of $70 \times 70$ pixels, named Patch, corresponding to $31\%$ of the original image. The second set contains images whose patch is $50 \times 50$ pixels, named Patch(S), covering $22\%$ of the original image (Figure \ref{fig:snpl_fgsm}). The reason for generating two sets with patches of different sizes was to assess whether the performance of the detection algorithms depends on the patch size, how much the results vary, or whether it ultimately depends on the pure disturbance regardless of size. We placed the patch randomly, without discriminating its location even when the patch was wholly or partially covering the object of interest.

We validated each dataset using the multipixel method, whose disturbance value $d = 10,000$ specifies the number of modified pixels. This result reflects a significant number of pixels that are not inconspicuous while allowing protruding objects to remain noticeable. This attack did not require knowledge of the internal parameters of any machine learning approach.

Next, we created for each database a set of AEs using Gaussian noise. Regarding all generated examples, we apply a $\sigma = 30$, which implements a grain of magnitude sufficient to distinguish the objects in the scene without a problem. Generating these examples did not require knowing the internal parameters of any learning approach. Also, we validated each dataset with Salt \& Pepper noise. Instead of applying it to grayscale images, we use color images, so the noise does not appear as black and white points but as color dots. This effect is because the Salt \& Pepper noise takes the values of the maximum and minimum pixels of each color band and distributes them throughout the image. The generation of Salt \& Pepper noise did not require knowledge of the internal parameters of any neural network. Finally, we create AEs for each dataset using Speckle noise. Since Speckle noise originates from physical conditions, it was necessary to implement a method to simulate it. Thus, random values were taken from a Gaussian distribution along image dimensions and multiplied by a variance of $0.3$. Similar to the other noises, the Speckle noise does not require knowing the parameters of any neural network to implement it.

\subsection{Validation Results}

In this section, we explain how the experiment demonstrates the threats that recent models could face. We provide four tables divided by each studied database: FT, ImgSal, PASCAL-S, and SNPL. We organize each table by the kind of attack: a white box attack (FGSM), the transferability effect (AP), and black box attacks (multipixel and noise-based attacks). Values in bold correspond to the lowest score an algorithm obtained among all types of AAs. Finally, we present the results of applying the symbolic and subsymbolic methods learned with the FT dataset to the AAs made on the DUTS database. The goal is to validate the robustness of the proposed models using a different dataset from that used during learning.

\subsubsection{FT database}

Table \ref{table:ft_results_fgsm} shows the results of applying an adversarial FGSM attack to the five neural networks and the BP algorithm using the FT database. In tests with the original images (no AA), neural networks outperformed the BP algorithm. However, as the attack is applied and the epsilon value increases, the networks' performance drops. PICANet's performance suffers a significant loss, starting with a score of $81.45\%$ and steadily falling to $40.52\%$ at $\epsilon = 64$, thus losing around $50\%$ of its performance. Although the BASNet network has the best score in the test, with the original images achieving $92.67\%$, it also loses its performance and drops to $58.65\%$. The second best model with the original images is EDN, which achieved $90.93\%$ before dropping to $53.43\%$ with $\epsilon = 64$. LeNo achieves the fourth-best score at $87.49\%$ while declining to $65.54\%$, which is lower than those of DHSNet and BP.
Although the DHSNet network does not lower its score so drastically, it is also affected. On the other hand, BP is virtually unaffected after applying all epsilon values, indicating that it possesses significant robustness and is quite resistant to this type of attack, being able to detect protruding objects in the scene even with the most intense disturbance, which is unlike neural networks. It is worth mentioning that the FT database consists of images whose objects of interest cover a large part of the image, so these algorithms have an excellent facility for achieving detection, which is reflected in the high scores.

Table \ref{table:ft_results_fgsm} also shows the results when applying the multipixel, AP, and noise attacks to the five neural networks and the BP algorithm using the FT database. PICANet is primarily affected by the multipixel attack, reaching the lowest score of $66.86\%$, while EDN and LeNo reached the second and third lowest scores, respectively. The remaining methods (BP, BASNet, and DHSNet) keep their scores after the multipixel attack. We can observe that the AP causes a score decrease in all detection algorithms, including BP. However, this decrease is more severe in the BASNet and PICANet networks, which had shown a high score with the undisturbed images. The DHSNet network, which had been shown to be unaffected by the FGSM attack, finally loses its score with the AP, dropping to $74.48\%$. On the other hand, the BP algorithm shows for the first time a decrease in its detection capacity, down to $67.96\%$, which suggests that the AP is a fairly powerful attack since the patches inserted in the images are, in many cases, classified as outstanding objects. EDN and LeNo scores also decreased after AP was applied. Lastly, although the smallest AP manages to lower the score of most algorithms, it fails to compare with the score of the largest patch, which was expected but may not be the case in the following experiments.

Regarding noise, the detection algorithms BP, BASNET, and DHSNet are practically unaffected, but not so with LeNo, EDN, and PICANet. These last three networks lose detection ability against all three types of noise, although the score loss is not as drastic, suggesting that they are still resistant to these attacks. It is important to note that one of the possible reasons for the high robustness to noise attacks is because this database, as already mentioned, contains objects of interest of considerable size and high contrast with the background, making them easy to spot. This feature contrasts with the results of the following databases, where noise attacks affect the score of the neural networks while BP maintains its high robustness.

\begin{table*}[!h]
\caption{\label{table:ft_results_fgsm} Summary of the results obtained with adversarial attacks (FGSM, multipixel, patch, and noises) applied to the FT database.}
\centering
\tiny
\begin{tabular}{ | l | l | l | l | l | l | l | l | l | l | l | l | l | l | l |}
 \hline
 \multicolumn{15}{|c|}{\textbf{FT}} \\
 \hline
 Algorithm & Measure & Original & $\epsilon=2$ & $\epsilon=4$ & $\epsilon=8$ & $\epsilon=16$ & $\epsilon=32$ & $\epsilon=64$ & Multipixel & Patch & Patch(S) & Gaussian & S\&P & Speckle \\
 \hline
 \multirow{2}{5em}{Brain \cite{Olague-SOD-2022} Programming} & Avg. & 73.84 & 73.79 & 73.63 & 73.69 & 73.44 & 72.97 & 71.64 & 74.31 & \textbf{67.96} & 69.63 & 73.84 & 73.90 & 73.96\\
 & $\sigma$ & 12.48 & 12.53 & 12.52 & 12.53 & 12.56 & 12.70 & 13.68 & 12.23 & 14.02 & 14.06 & 12.28 & 12.60 & 12.35 \\
 \hline
 \multirow{2}{5em}{LeNo \cite{Wang2023}} & Avg. & 87.49 & 87.40 & 87.17 & 87.16 & 86.74 & 83.65 & \textbf{65.54} & 72.99 & 80.17 & 80.61 & 86.59 & 83.56 & 82.37\\
 & $\sigma$ & 14.38 & 14.43 & 14.85 & 14.58 & 15.18 & 17.57 & 25.22 & 24.17 & 16.31 & 16.34 & 15.55 & 18.64 & 19.35 \\
 \hline
 \multirow{2}{5em}{EDN \cite{Wu2022}} & Avg. & 90.93 & 90.87 & 90.77 & 90.42 & 88.25 & 82.20 & \textbf{53.43} & 69.10 & 81.48 & 83.28 & 86.08 & 86.01 & 82.88\\
 & $\sigma$ & 12.82 & 12.81 & 12.89 & 13.12 & 16.12 & 21.04 & 26.38 & 26.95 & 17.06 & 16.12 & 18.33 & 18.77 & 21.95\\
 \hline
 \multirow{2}{5em}{BASNet \cite{qin_basnet_2021}} & Avg. & 92.67 & 90.92 & 89.74 & 88.92 & 87.30 & 82.60 & \textbf{58.65} & 90.79 & 67.01 & 73.00 & 92.27 & 92.41 & 90.88\\
 & $\sigma$ & 12.51 & 15.38 & 16.94 & 16.78 & 18.02 & 23.28 & 37.96 & 15.54 & 20.47 & 23.16 & 12.72 & 12.69 & 14.75 \\
 \hline
 \multirow{2}{5em}{PICANet \cite{liu_picanet_2018}} & Avg. & 81.45 & 74.26 & 67.84 & 61.01 & 57.04 & 51.21 & \textbf{40.52} & 66.86 & 61.04 & 69.78 & 78.39 & 78.95 & 75.89\\
 & $\sigma$ & 15.27 & 18.74 & 20.83 & 22.58 & 22.91 & 22.76 & 22.19 & 22.14 & 19.79 & 17.06 & 17.42 & 19.40 & 20.97 \\
 \hline
 \multirow{2}{5em}{DHSNet \cite{liu_dhsnet_2016}} & Avg. & 88.63 & 87.92 & 87.46 & 86.68 & 85.62 & 83.85 & 78.92 & 85.67 & \textbf{74.48} & 79.04 & 87.84 & 86.98 & 86.89\\
 & $\sigma$ & 12.43 & 12.86 & 13.30 & 13.74 & 14.48 & 15.95 & 19.90 & 14.50 & 17.67 & 16.38 & 13.31 & 14.12 & 14.73\\
 \hline
\end{tabular}
\end{table*}

\subsubsection{ImgSal database}
Table \ref{table:imgsal_results_fgsm} shows the results of applying FGSM to the five neural networks and the BP algorithm using the ImgSal database. It is important to emphasize that the ImgSal database does not have correctly segmented ground truths, so the results in the table should be taken as an approximation of the actual score that the algorithms would have returned if the segmentation had been correct. In this case, the five neural networks, even BASNet and PICANet, are severely affected by the FGSM attack, losing great detection capacity from the value $\epsilon = 8$. Their performance has already dropped considerably for the value $\epsilon= 64$, in the following order: EDN, BASNet, PICANet, LeNo, and DHSNet, down to $25\%$ for EDN. BP remains at an acceptable score even after the intense disturbance $\epsilon = 64$, dropping only by $8\%$. This result again underscores the robustness of BP against this type of AA.

Table \ref{table:imgsal_results_fgsm} also displays the results when applying the multipixel, AP, and noise attacks to the five neural networks and the BP algorithm using the ImgSal database. In this case, unlike the FT databases, the multipixel attack significantly affects the five neural networks' performance. This result may be because the objects in the ImgSal database images are not as large as those in FT, so detection is more challenging. The BP algorithm is not affected by this attack. Regarding the AP, this seriously affects the detection capacity of all algorithms, with BASNet and PICANet losing around 50\% of their performance. Also, the small patch causes singular behavior in the LeNo, BP, and DHSNet  algorithms; the score obtained is lower than in the regular-sized patch. This result indicates that a larger patch will not necessarily affect the performance of an algorithm to a greater extent.

In the case of noise, BP is not affected by any method, keeping its score very close to the original. Instead, the five neural networks decreased their performance, especially in Salt \& Pepper and Speckle noises. Unlike FT, where no algorithm was affected by noise, the case of ImgSal does affect them due to the nature of its images, which pose an even more significant challenge at the time of detection.

\begin{table*}[!h]
\caption{\label{table:imgsal_results_fgsm} Summary of the results obtained with adversarial attacks (FGSM, multipixel, patch, and noises) applied to the ImgSal database.}
\centering
\tiny
\begin{tabular}{ | l | l | l | l | l | l | l | l | l | l | l | l | l | l | l |}
 \hline
 \multicolumn{15}{|c|}{\textbf{ImgSal}} \\
 \hline
 Algorithm & Measure & Original & $\epsilon=2$ & $\epsilon=4$ & $\epsilon=8$ & $\epsilon=16$ & $\epsilon=32$ & $\epsilon=64$ & Multipixel & Patch & Patch(S) & Gaussian & S\&P & Speckle\\
 \hline
 \multirow{2}{5em}{Brain \cite{Olague-SOD-2022} Programming} & Avg. & 62.66 & 62.56 & 62.52 & 61.96 & 60.30 & 58.87 & 54.20 & 60.37 & 46.25 & \textbf{46.12} & 62.27 & 62.18 & 61.28\\
 & $\sigma$ & 23.96 & 23.83 & 24.09 & 24.11 & 24.43 & 24.72 & 24.73 & 23.54 & 25.13 & 24.61 & 23.91 & 24.21 & 23.73\\
 \hline
 \multirow{2}{5em}{LeNo \cite{Wang2023}} & Avg. & 72.80 & 72.54 & 72.60 & 72.74 & 70.75 & 57.62 & \textbf{28.92} & 60.79 & 59.51 & 57.42 & 69.73 & 67.42 & 63.80\\
 & $\sigma$ & 24.52 & 25.10 & 24.70 & 24.48 & 25.80 & 29.76 & 24.15 & 29.48 & 26.80 & 27.38 & 25.88 & 24.69 &  27.37\\
 \hline
 \multirow{2}{5em}{EDN \cite{Wu2022}} & Avg. & 69.85 & 69.30 & 69.00 & 69.09 & 63.63 & 42.62 & \textbf{17.41} & 41.27 & 52.17 & 54.55 & 55.28 & 53.95 & 44.66\\
 & $\sigma$ & 24.73 & 25.22 & 24.98 & 24.64 & 27.07 & 30.24 & 16.72 & 30.25 & 28.22 & 27.92 & 30.22 & 29.76 & 29.98\\
 \hline
 \multirow{2}{5em}{BASNet \cite{qin_basnet_2021}} & Avg. & 63.75 & 59.84 & 55.96 & 53.78 & 54.76 & 44.34 & \textbf{20.69} & 54.19 & 27.92 & 31.49 & 60.01 & 57.27 & 55.06 \\
 & $\sigma$ & 31.73 & 33.32 & 34.43 & 35.62 & 34.19 & 35.76 & 30.10 & 35.81 & 26.28 & 27.64 & 34.68 & 34.57 & 34.18\\
 \hline
 \multirow{2}{5em}{PICANet \cite{liu_picanet_2018}} & Avg. & 71.60 & 63.73 & 56.14 & 50.71 & 49.52 & 44.85 & \textbf{25.78} & 50.33 & 35.6 & 38.49 & 65.12 & 67.23 & 64.29\\
 & $\sigma$ & 22.16 & 28.51 & 29.29 & 28.36 & 28.58 & 28.75 & 22.96  & 30.62 & 23.57 & 22.48 & 29.82 & 30.56 & 29.41\\
 \hline
 \multirow{2}{5em}{DHSNet \cite{liu_dhsnet_2016}} & Avg. & 53.63 & 53.25 & 52.51 & 50.86 & 49.22 & 45.96 & \textbf{39.89} & 49.12 & 41.94 & 41.60 & 51.19 & 48.83 & 49.04 \\
 & $\sigma$ & 24.85 & 24.62 & 24.73 & 24.97 & 24.81 & 24.98 & 24.34 & 25.66 & 26.80 & 25.84 & 24.83 & 25.37 & 25.77 \\
 \hline
\end{tabular}
\end{table*}

\subsubsection{PASCAL-S database}
Table \ref{table:pascal_results_fgsm} shows the results of applying an adversarial FGSM attack to the five neural networks and the BP algorithm using the PASCAL database. At first glance, we can appreciate that the BP algorithm is not affected by this attack, keeping its score very close to the original and consistently above $60\%$. This result demonstrates once again the excellent robustness of BP. On the other hand, as we already observed in previous databases, neural networks do not keep up with this type of attack, and this case is no exception. Even though BASNet had the highest score in the original images ($80.20\%$), its score dropped to $14.96\%$ in the highest epsilon. The BP algorithm may not have had the highest score using the original images, but it outperforms four out of five neural networks, with the exception of LeNo, when applying the FGSM with the most severe perturbation.

Table \ref{table:pascal_results_fgsm} also shows the results when applying the multipixel, AP, and noise attacks to the five neural networks and the BP algorithm using the PASCAL-S database. In the case of the multipixel attack, the neural networks are considerably affected, especially BASNet, which drops from $80.20\%$ to $57.26\%$. For the AP, the BASNet score drops even further, while PICANet mysteriously recovers at $4.63\%$ compared with its multipixel score. This result indicates that the algorithm's performance highly correlates with the database, and in the AP, the PICANet algorithm tolerated the patch slightly better than EDN. Note that the LeNo score is slightly better than the PICANet score regarding the AP. However, robustness is more evident in the BP algorithm, which can acceptably withstand these two types of attacks, starting at $62.94\%$ in the original images and up to $57.5\%$ in Patch(S). Again, attacking BP with a normal-sized patch did not result in a lower score than attacking it with a small patch.

In the case of noise attacks, the BP and LeNo algorithms are largely unaffected, but we cannot say the same regarding the other neural networks. Even though it seems that the noise attack does not manage to affect the performance of these four neural networks so drastically, there is no doubt that there is a decrease in the detection capacity. EDN, BASNet, and PICANet are down an average of 
$8\%$\textendash$15\%$, while DHSNet, which already proves robust to noise attacks, is down only 3\% this time.

\begin{table*}[!h]
\caption{\label{table:pascal_results_fgsm} Summary of the results obtained with adversarial attacks (FGSM, multipixel, patch, and noises) applied to the PASCAL-S database.}
\centering
\tiny
\begin{tabular}{ | l | l | l | l | l | l | l | l | l | l | l | l | l | l | l | }
 \hline
 \multicolumn{15}{|c|}{\textbf{PASCAL-S}} \\
 \hline
 Algorithm & Measure & Original & $\epsilon=2$ & $\epsilon=4$ & $\epsilon=8$ & $\epsilon=16$ & $\epsilon=32$ & $\epsilon=64$ & Multipixel & Patch & Patch(S) & Gaussian & S\&P & Speckle \\
 \hline
 \multirow{2}{5em}{Brain \cite{Olague-SOD-2022} Programming} & Avg. & 62.94 & 62.79 & 62.34 & 62.53 & 62.38 & 61.53 & 60.53 & 61.74 & 58.05 & \textbf{57.50} & 62.55 & 61.51 & 61.55\\
 & $\sigma$ & 20.89 & 20.65 & 20.80 & 21.13 & 21.60 & 21.28 & 21.54 & 18.28 & 18.84 & 17.96 & 18.12 & 18.41 & 18.58\\
 \hline
 \multirow{2}{5em}{LeNo \cite{Wang2023}} & Avg. & 73.59 & 73.11 & 73.48 & 72.97 & 72.01 & 70.01 & \textbf{62.29} & 69.65 & 69.74 & 70.37 & 72.58 & 72.51 & 72.72\\
 & $\sigma$ & 17.15 & 17.16 & 17.19 & 17.27 & 17.62 & 18.56 & 21.33 & 19.51 & 18.49 & 17.54 & 17.94 & 17.45 & 17.25\\
 \hline
 \multirow{2}{5em}{EDN \cite{Wu2022}} & Avg. & 72.68 & 72.49 & 72.24 & 71.82 & 69.07 & 58.11 & \textbf{47.96} & 57.45 & 66.52 & 68.89 & 65.26 & 67.26 & 63.93\\
 & $\sigma$ & 18.67 & 18.78 & 18.99 & 18.83 & 19.45 & 22.13 & 21.99 & 21.85 & 20.50 & 18.84 & 20.91 & 19.71 & 21.29\\
 \hline
 \multirow{2}{5em}{BASNet \cite{qin_basnet_2021}} & Avg. & 80.20 & 77.24 & 74.64 & 71.13 & 65.96 & 50.82 & \textbf{14.96} & 57.26 & 53.12 & 54.67 & 67.66 & 74.66 & 66.49\\
 & $\sigma$ & 21.15 & 23.16 & 23.53 & 26.03 & 27.74 & 33.78 & 29.49 & 32.54 & 26.26 & 29.63 & 28.86 & 25.69 & 29.32\\
 \hline
 \multirow{2}{5em}{PICANet \cite{liu_picanet_2018}} & Avg. & 75.81 & 68.16 & 59.38 & 51.12 & 52.00 & 53.61 & \textbf{50.08} & 63.31 & 67.94 & 69.34 & 70.16 & 72.06 & 68.65 \\
 & $\sigma$ & 19.80 & 20.74 & 20.17 & 21.48 & 21.54 & 20.64 & 20.33 & 19.69 & 17.24 & 17.53 & 17.29 & 17.25 & 19.19 \\
 \hline
 \multirow{2}{5em}{DHSNet \cite{liu_dhsnet_2016}} & Avg. & 68.14 & 66.57 & 65.41 & 63.44 & 61.65 & 60.03 & \textbf{57.07} & 64.54 & 62.03 & 64.41 & 65.69 & 66.32 & 65.19\\
 & $\sigma$ & 21.03 & 21.36 & 21.07 & 20.96 & 21.17 & 21.37 & 21.38 & 20.92 & 21.67 & 20.83 & 20.46 & 20.97 & 21.42\\
 \hline
\end{tabular}
\end{table*}

\subsubsection{SNPL database}

Unlike previous databases, the Snowy Plover (SNPL) images contain characteristics that come quite close to a real-world problem. In other words, we planned the photographs of these images in natural environments, with cameras for bird photography, and with different weather, lighting, and zoom conditions. This dataset allows it to test detection algorithms that are mostly only tested against databases with larger objects of interest that are optimally illuminated and have no real-world purpose.

Table \ref{table:snpl_results_fgsm} shows the results of applying an adversarial FGSM attack to the five neural networks and the BP algorithm using the SNPL database. The first observation in this experiment is the decrease in the scores of most of the algorithms compared with the previous databases. Except for PICANet in the test with the original images, the neural networks and BP lose most of their performance when trying to achieve detection with the SNPL database. This result is due to the different zoom and obstruction characteristics in this set of images. Even though the neural networks scored reasonably well against the other databases, where the objects of interest have a relatively high salience, a more significant challenge begins to break them. This outcome becomes much more evident when attacking them with AEs. For the case of FGSM, all neural networks show a gradual decrease starting at $\epsilon = 2$ and ending with a practically nonexistent score when attacking with a value of $\epsilon = 64$. LeNo loses more than half of the score reaching $39.79\%$, while BASNet completely loses the score with $0.99\%$. In the case of BP, when testing the BP algorithm with the original images, it starts with a score of $52.6\%$, below those of the PICANet, LeNo, EDN, and BASNet scores, in order of performance, but still better than that of DHSNet. However, by attacking with $\epsilon = 32$, BP already beat all neural networks, finishing with a score well above theirs. In the end, when attacking with $\epsilon = 64$, the BASNet network shows a total decrease of $98.3\%$, $89.6\%$ for PICANet, $87.8\%$ for EDN, $59.4\%$ for DHSNet, and $55.86\%$ for LeNo, while BP only drops to $19.7\%$.

Table \ref{table:snpl_results_fgsm} also shows the results when applying the multipixel, AP, and noise attacks to the five neural networks and the BP algorithm using the SNPL database. In the case of the multipixel attack, the massive decrease in the score of PICANet and BASNet is remarkable, to the point that the PICANet network dropped from $86.04\%$ to $13.29\%$ only with this attack, and the BASNet network suffered the same consequences. Also, EDN and LeNo achieved poor performances, decreasing their scores by $46.44\%$ and $38.06\%$, respectively. Again, due to the nature of the SNPL database images, these networks fail to maintain the robustness necessary to achieve detection. The BP algorithm only shows a 9.3\% decrease in its score, demonstrating its high robustness. For the case of the AP, PICANet manages to be surprisingly robust, going down from $86.04\%$ to $72.06\%$ in the regular-size patch and $82.66\%$ in the smallest size patch, achieving a high level of detection even when we designed the patch to be another type of protruding object in the scene. On the other hand, BP manages to hold close to $40\%$ of the patch attacks of both sizes, outperforming the BASNet and DHSNet networks. Note that BASNet scored the worst results, with $20.47\%$ for the regular-size patch and $20.53\%$ for the small patch. LeNo decreases its score reaching to $53.17\%$ for the regular patch while improving up to $61.08\%$ for the small patch, losing $24\%$ and $12.7\%$, respectively. EDN maintains a similar score for both patches at around $49\%$, but its performance still decreased by $25\%$.

Noise attacks continue to show that neural networks are not immune to them. In all types of noise attacks, the BP algorithm outperforms the five neural networks, with only Speckle noise managing to lower its score to $44.98\%$ while remaining above all networks. PICANet and BASNet suffer huge performance drops, with Speckle noise destroying their scores almost entirely (PICANet down to $91\%$ and BASNet down to a total of $80\%$). LeNo, DHSNet, and EDN demonstrate some robustness, but not enough to outperform BP.

\begin{table*}[!h]
\caption{\label{table:snpl_results_fgsm} Summary of the results obtained with adversarial attacks (FGSM, multipixel, patch, and noises) applied to the SNPL database.}
\centering
\tiny
\begin{tabular}{ | l | l | l | l | l | l | l | l | l | l | l | l | l | l | l | }
 \hline
 \multicolumn{15}{|c|}{\textbf{SNPL}} \\
 \hline
 Algorithm & Measure & Original & $\epsilon=2$ & $\epsilon=4$ & $\epsilon=8$ & $\epsilon=16$ & $\epsilon=32$ & $\epsilon=64$ & Multipixel & Patch & Patch(S) & Gaussian & S\&P & Speckle \\
 \hline
 \multirow{2}{5em}{Brain \cite{Olague-SOD-2022} Programming} & Avg. & 52.60 & 53.31 & 52.13 & 50.31 & 49.09 & 45.01 & 42.24 & 47.70 & 41.26 & \textbf{39.32} & 48.40 & 45.74 & 44.98 \\
 & $\sigma$ & 19.72 & 19.23 & 19.64 & 20.73 & 21.21 & 20.27 & 20.28 & 21.12 & 21.65 & 19.89 & 20.98 & 21.91 & 21.26\\
 \hline
 \multirow{2}{5em}{LeNo \cite{Wang2023}} & Avg. & 69.97 & 69.72 & 70.51 & 69.42 & 63.80 & 39.79 & \textbf{30.88} & 43.34 & 53.17 & 61.08 & 51.39 & 46.58 & 39.15\\
 & $\sigma$ & 17.67 & 17.49 & 16.90 & 17.66 & 19.21 & 21.98 & 20.23 & 23.98 & 19.95 & 20.41 & 22.03 & 25.01 & 21.92\\
 \hline
 \multirow{2}{5em}{EDN \cite{Wu2022}} & Avg. & 66.32 & 66.10 & 65.44 & 62.78 & 52.22 & 32.69 & \textbf{8.10} & 35.52 & 49.43 & 49.58 & 42.46 & 42.75 & 30.95\\
 & $\sigma$ & 18.23 & 18.64 & 19.09 & 20.95 & 23.87 & 23.99 & 6.54 & 26.22 & 21.34 & 24.74 & 24.73 & 27.08 & 23.83\\
 \hline
 \multirow{2}{5em}{BASNet \cite{qin_basnet_2021}} & Avg. & 60.27 & 57.36 & 53.88 & 50.66 & 35.28 & 13.63 & \textbf{0.99} & 18.09 & 20.47 & 20.53 & 31.27 & 30.63 & 10.01 \\
 & $\sigma$ & 35.12 & 35.40 & 34.43 & 31.85 & 31.80 & 24.72 & 2.05 & 28.11 & 21.30 & 26.71 & 34.50 & 34.73 & 19.84 \\
 \hline
 \multirow{2}{5em}{PICANet \cite{liu_picanet_2018}} & Avg. & 86.04 & 72.91 & 63.7 & 57.64 & 44.85 & 9.26 & \textbf{8.90} & 13.29 & 72.61 & 82.66 & 32.79 & 26.93 & 7.37 \\
 & $\sigma$ & 14.64 & 23.56 & 25.12 & 25.32 & 22.41 & 8.16 & 5.65 & 12.09 & 25.58 & 21.02 & 22.22 & 23.42 & 6.63 \\
 \hline
 \multirow{2}{5em}{DHSNet \cite{liu_dhsnet_2016}} & Avg. & 43.98 & 43.23 & 43.04 & 42.12 & 38.63 & 31.67 & \textbf{17.84} & 36.13 & 34.76 & 30.01 & 38.92 & 38.19 & 33.60\\
 & $\sigma$ & 19.88 & 19.46 & 19.37 & 19.88 & 19.57 & 19.48 & 15.77 & 21.02 & 17.76 & 15.59 & 19.42 & 20.86 & 20.01 \\
 \hline
\end{tabular}
\end{table*}

\subsubsection{DUTS database}

As a way to validate our results, we selected 300 random images from the 5019 testing pictures available in the DUTS dataset \cite{Wang2017}. In this manner, we were able to generate a validation set that was comparable in size to prior databases. Additionally, we used the models that resulted from training on the FT database so that we could observe how the different algorithms would behave on an unrelated collection of photos.

Table \ref{table:duts_results_fgsm} shows the results of applying an adversarial FGSM attack to the five neural networks and the BP algorithm using the DUTS database. The consistency in the results obtained by the BP algorithm is worth mentioning, hovering around 50\%. Similar to the results obtained with previous databases, the LeNo and EDN neural networks show resiliency for smaller gradient attacks, and then considerably deteriorate after $\epsilon = 32$ attacks. Scoring $57.1\%$ in the original images, LeNo drops its score to $37.0\%$ with $\epsilon = 64$, while EDN has a more prominent drop in score, starting at $58.63\%$, and falling down to $26.2\%$. BASNet has the highest score in this experiment, with $82.43\%$, but also starts deteriorating sooner with $\epsilon = 4$ attacks. PiCANet scores slightly better ($62.25\%$) than LeNo and EDN, but quickly deteriorates after $\epsilon = 2$ attacks, reaching a score of $19.37\%$ on the strongest epsilon attack. Finally, while DHSNet does not deteriorate as prominently as other neural networks, it does have the lowest score out of all algorithms, starting at $25.77\%$, and reaching a score of $16.42\%$ with $\epsilon = 64$.

Table \ref{table:duts_results_fgsm} also shows the results when applying the multipixel, AP, and noise attacks to the five neural networks and the BP algorithm using the DUTS database. Again, BP shows the highest robustness, with the AP causing its most significant drop in performance, going from $50.22\%$ down to $47.66\%$. PiCANet, EDN, LeNo, and DHSNet show a high sensitivity against multipixel, and speckle noise attacks, while algorithms such as BASNet seem to be most affected by AP attacks.

\begin{table*}[!h]
\caption{\label{table:duts_results_fgsm} Summary of the results obtained with adversarial attacks (FGSM, multipixel, patch, and noises) applied to the DUTS database, using models trained on the FT database.}
\centering
\tiny
\begin{tabular}{ | l | l | l | l | l | l | l | l | l | l | l | l | l | l | l | }
 \hline
 \multicolumn{15}{|c|}{\textbf{DUTS}} \\
 \hline
 Algorithm & Measure & Original & $\epsilon=2$ & $\epsilon=4$ & $\epsilon=8$ & $\epsilon=16$ & $\epsilon=32$ & $\epsilon=64$ & Multipixel & Patch & Patch(S) & Gaussian & S\&P & Speckle \\
 \hline
 \multirow{2}{5em}{Brain \cite{Olague-SOD-2022} Programming} & Avg. & 50.22 & 50.06 & 50.08 & 49.96 & 49.90 & 50.4 & 49.39 & 49.82 & \textbf{47.66} & 49.02 & 50.07 & 50.13 & 50.47\\
 & $\sigma$ & 22.23 & 22.32 & 22.33 & 22.24 & 22.42 & 22.63 & 22.40 & 22.47 & 22.25 & 22.39 & 22.27 & 22.36 & 22.37\\
 \hline
 \multirow{2}{5em}{LeNo \cite{Wang2023}} & Avg. & 57.10 & 57.59 & 57.70 & 56.77 & 55.13 & 49.13 & \textbf{37.00} & 47.21 & 51.96 & 51.03 & 54.10 & 50.57 & 48.16\\
 & $\sigma$ & 25.04 & 24.79 & 24.77 & 24.64 & 24.92 & 25.19 & 23.29 & 26.28 & 24.04 & 24.69 & 25.02 & 25.34 & 25.23\\
 \hline
 \multirow{2}{5em}{EDN \cite{Wu2022}} & Avg. & 58.63 & 58.53 & 58.32 & 57.71 & 55.40 & 44.02 & \textbf{26.20} & 44.70 & 50.73 & 52.28 & 52.25 & 51.09 & 46.51\\
 & $\sigma$ & 25.61 & 26.64 & 25.68 & 26.11 & 26.46 & 26.89 & 21.84 & 27.11 & 25.85 & 25.28 & 26.67 & 26.59 & 27.44\\
 \hline
 \multirow{2}{5em}{BASNet \cite{qin_basnet_2021}} & Avg. & 82.43 & 81.20 & 76.67 & 67.08 & 53.96 & 42.14 & \textbf{29.11} & 63.34 & 45.97 & 55.97 & 71.18 & 75.16 & 67.64\\
 & $\sigma$ & 24.13 & 24.87 & 28.10 & 31.97 & 33.47 & 30.92 & 23.10 & 30.47 & 26.73 & 29.69 & 28.81 & 27.14 & 30.33\\
 \hline
 \multirow{2}{5em}{PICANet \cite{liu_picanet_2018}} & Avg. & 62.25 & 55.86 & 42.06 & 31.59 & 27.48 & 24.17 & \textbf{19.37} & 40.62 & 47.71 & 48.29 & 50.61 & 51.06 & 47.34\\
 & $\sigma$ & 23.64 & 24.81 & 24.55 & 22.15 & 20.75 & 19.33 & 16.64 & 23.87 & 24.34 & 23.63 & 24.94 & 26.30 & 25.79\\
 \hline
 \multirow{2}{5em}{DHSNet \cite{liu_dhsnet_2016}} & Avg. & 25.77 & 25.71 & 25.54 & 24.68 & 21.65 & 17.32 & \textbf{16.42} & 17.25 & 23.90 & 24.50 & 18.48 & 19.90 & 18.16\\
 & $\sigma$ & 17.81 & 17.79 & 17.71 & 17.36 & 16.17 & 14.64 & 14.59 & 14.70 & 16.76 & 17.16 & 14.84 & 15.43 & 14.93\\
 \hline
\end{tabular}
\end{table*}

\begin{figure}[!h]
    \centering
    \includegraphics[width=1.0\linewidth]{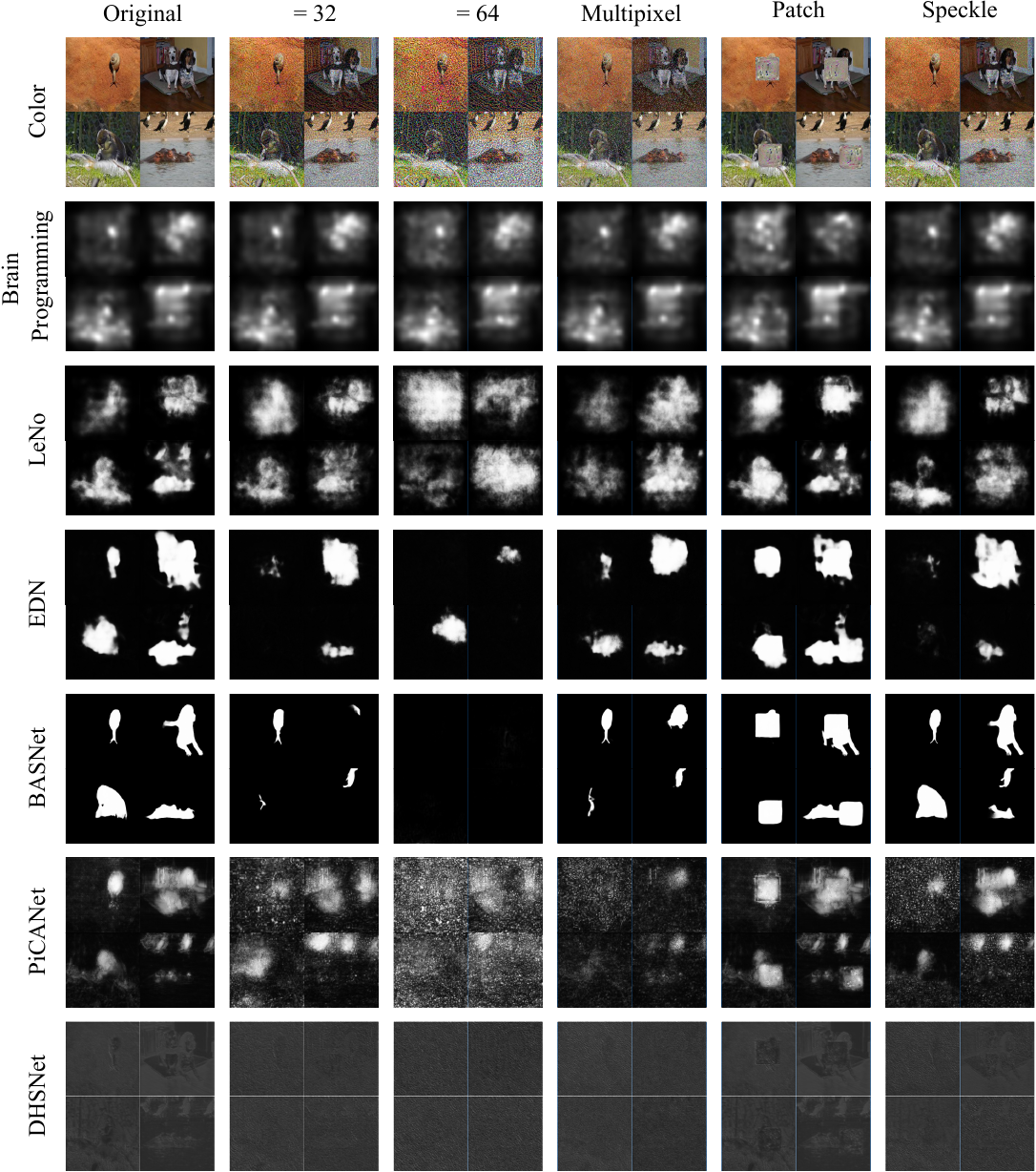}
    \setlength{\belowcaptionskip}{10pt}
    \caption{Saliency maps are generated on the DUTS database using algorithms trained with the FT database. Columns group results by the original image, FGSM attack with $\epsilon=32$ and $\epsilon=64$, multipixel, patch, and speckle noise attacks. The first row shows the original colored image with its corresponding AAs, followed by the results generated by the various algorithms.}
    \label{fig:duts_results_proto_objects}
\end{figure}

Figure \ref{fig:duts_results_proto_objects} illustrates example saliency maps generated on the validation DUTS dataset and provides a visual explanation of the behavior observed in Table \ref{table:duts_results_fgsm}. We included AAs that produced the most variance across all algorithms in comparison with the original saliency maps. For reference, we also included the colored images that were passed as inputs to the various algorithms to achieve a better understanding of the produced outputs. Starting with BP, the first thing that comes to our attention is how constant the resulting saliency maps are. The cases where we can see the most distortion are in the AP, just as the table results hint. If we look carefully at the colored images, we will notice that the cases showing the biggest distortion are those where the patch fully covers the salient object. Now, the LeNo algorithm produces results that are reminiscent of smoke; it is worth noting how every single one of the attacks portrayed in the figure manages to completely change the computed output. While LeNo showed the highest resiliency out of the five networks according to the table results, these examples show that even the most robust state-of-the-art neural network does not comply with the proposed definition of robustness. In the cases of EDN and BASNet, we can observe how FGSM attacks tend to obscure the resulting saliency maps, leading to the highest loss in performance. PiCANet seems to be very susceptible to various attacks since the algorithm appears to prefer the attacked pixels of the image over the salient object. Lastly, we can observe how DHSNet is also affected by AAs, failing to produce consistent saliency maps.

\subsubsection{Summary of Results}

Table \ref{table:std} shows the standard deviations of each of the detection models for each database. Here, it is clearly shown that the dispersion of the results obtained by the BP algorithm is low compared with those of all other neural networks. This outcome is due to the high robustness of BP, which keeps the experiments' values with AAs close to the original images. In the case of neural networks, the results deviated quite a bit, especially in the case of PICANet for the SNPL database. Even though it scored highly in the original images, it fell apart in the attacks. DHSNet also demonstrated low dispersion in the data; however, as seen in the tables above, it is susceptible to attacks such as FGSM, so it cannot compete with BP.

\begin{table}[!h]
\caption{\label{table:std} Standard deviation of each detection model with respect to adversary attack experiments for each database.}
\centering
\begin{tabular}{ | p{1.8cm} | l | l | l | l | l | }
 \hline
 \multicolumn{6}{|c|}{\textbf{Standard Deviation $\sigma$}} \\
 \hline
 Algorithm & FT & ImgSal & PASCAL-S & SNPL & DUTS \\
 \hline
 BP  \cite{Olague-SOD-2022} & 1.93 & 5.96 & 1.73 & 4.47 & 0.75\\
 \hline
 LeNo \cite{Wang2023} & 6.56 & 12.02 & 3.03 & 13.81 & 5.77\\
 \hline
 EDN \cite{Wu2022} & 10.63 & 15.15 & 7.32 & 17.10 & 8.93\\
 \hline
 BASNet \cite{qin_basnet_2021} & 11.09 & 13.74 & 17.14 & 19.39 & 16.20\\
 \hline
 PICANet \cite{liu_picanet_2018} & 12.04 & 13.82 & 8.90 & 29.76 & 12.94\\
 \hline
 DHSNet \cite{liu_dhsnet_2016} & 4.37 & 4.56 & 3.02 & 7.15 & 3.67\\
 \hline
\end{tabular}
\end{table}

\section{Conclusion}

In this work, we studied SOD with an emphasis on measuring the robustness of various algorithms concerning AAs. The BP methodology was defined to detect conspicuous objects by involving a process similar to that of the artificial visual cortex in humans. After the experiments, BP proved highly robust, supporting even the highest-disturbance images. AEs were implemented, such as the FGSM, AP, Multipixel Attack, and three types of noise: Gaussian, Salt \& Pepper, and Speckle. By attacking the neural networks with each of these AEs, we confirmed that they are not reliable at all. Even the slightest disturbance in the input images lowers their performance significantly, not to mention that feeding them a strong disturbance renders them virtually unusable. The advantage of BP is that it is highly robust and reliable during detection, and although a neural network achieves higher performances than the evolutionary algorithm, it can collapse with virtually any disturbance to its input data. These experiments were carried out in five databases: FT, ImgSal, PASCAL, DUTS, and SNPL. The first four contain highly detectable and unnatural objects since they cover a large part of the scene and have high contrast. Nevertheless, the SNPL database is different, as it contains objects with characteristics that resemble the real world more closely. By using this database, we can get closer to a detection problem with a practical purpose that directly helps solve a real-world problem, which is, in this case, wildlife conservation. In future research, we will focus on improving the performance of BP.

\ifCLASSOPTIONcompsoc
  \section*{Acknowledgments}
\else
  \section*{Acknowledgment}
\fi

The following projects support this research:
1) Project titled ``Estudio de la programaci\'{o}n cerebral en problemas de reconocimiento a gran escala y sus aplicaciones en el mundo real'' CICESE-634135. 
2) Project titled ``Deep learning applications for computer vision task'' funded by NITROAA with support of Lenovo P920 and Dell Inception 7820 workstation and NVIDIA Corporation with support from the NVIDIA Titan V and Quadro RTX 8000 GPU.
3) Project titled ``Applications of Drone Vision using Deep Learning' funded by Technical Education Quality Improvement Programme (referred to as TEQIP-III), National Project Implementation Unit, Government of India.
Jonathan Vargas graciously acknowledges the economic support granted by Cornell University through the program of Coastal Solutions Fellows Program. Roberto Pineda would like to thank CONACyT Mexico for the scholarship awarded to conduct studies at CICESE. Matthieu Olague would like to thank the An\'{a}huac University--Quer\'{e}taro for the scholarship granted to carry out his studies in Mechatronics Engineering.

\ifCLASSOPTIONcaptionsoff
  \newpage
\fi

\begin{IEEEbiography}[{\includegraphics[width=1in,height=1.25in,clip,keepaspectratio]{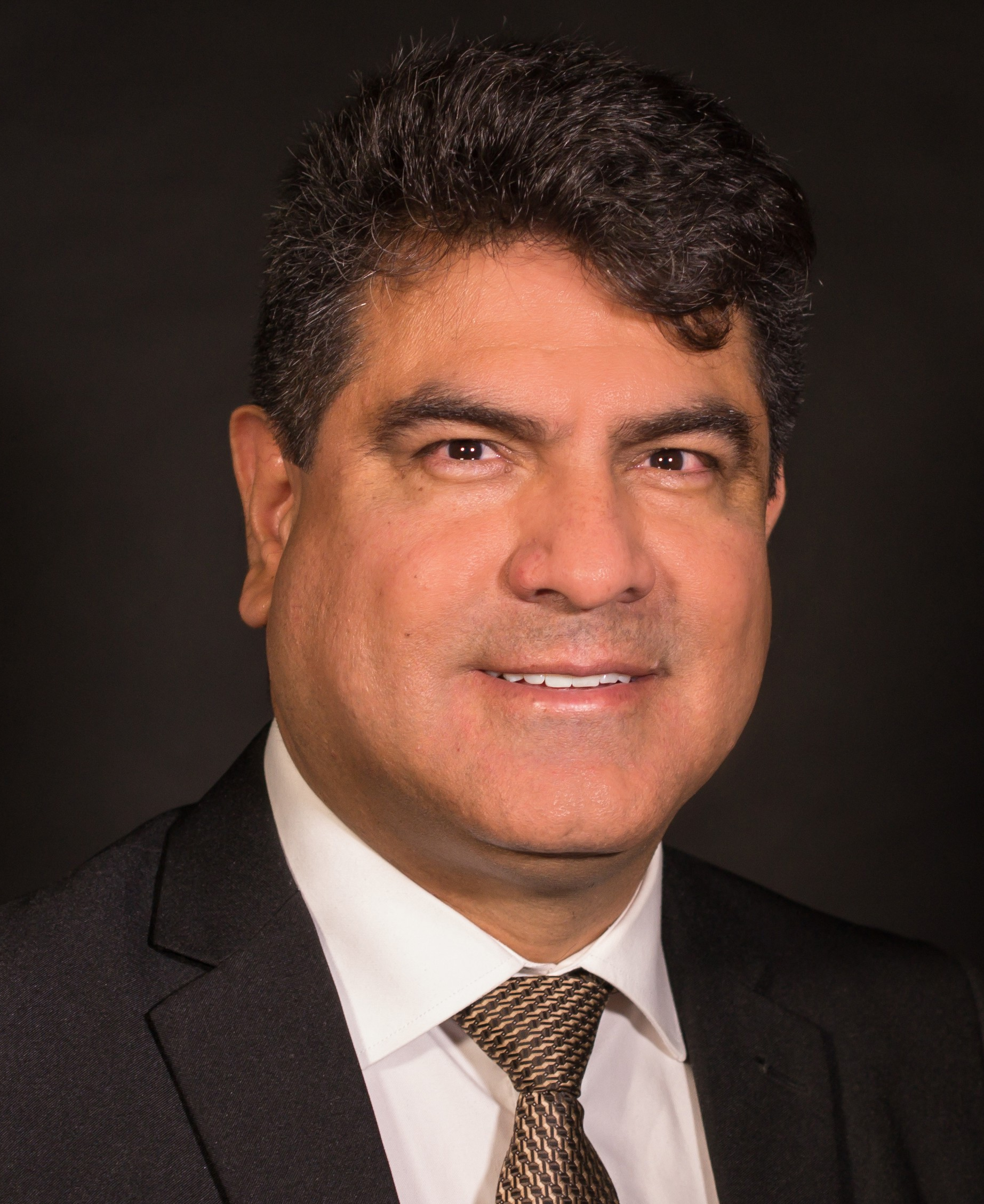}}]{Gustavo Olague} (M'99--SM'17) received the B.S. and M.S. degrees in Industrial and Electronics Engineering from ITCH (Instituto Tecnol\'{o}gico de Chihuahua) in 1992 and 1995 respectively, and his Ph.D. in Computer Vision, Graphics, and Robotics from INPG and INRIA in France in 1998. He is a Professor in the Dept. of Computer Science at CICESE in M\'{e}xico. He has authored over 150 conference proceedings papers and journal articles and co-edited special issues in Pattern Recognition Letters, Evolutionary Computation, and Applied Optics. Prof. Olague received numerous distinctions, among them the Talbert Abrams Award presented by the American Society for Photogrammetry and Remote Sensing (ASPRS); Outstanding Associate Editor IEEE Access 2021; Best Paper Awards at major conferences; and twice the Bronze Medal at the Humies (GECCO award). Gustavo Olague is the author of the book Evolutionary Computer Vision published by Springer in the Natural Computing Series.
\end{IEEEbiography}

\begin{IEEEbiography}[{\includegraphics[width=1in,height=1.25in,clip,keepaspectratio]{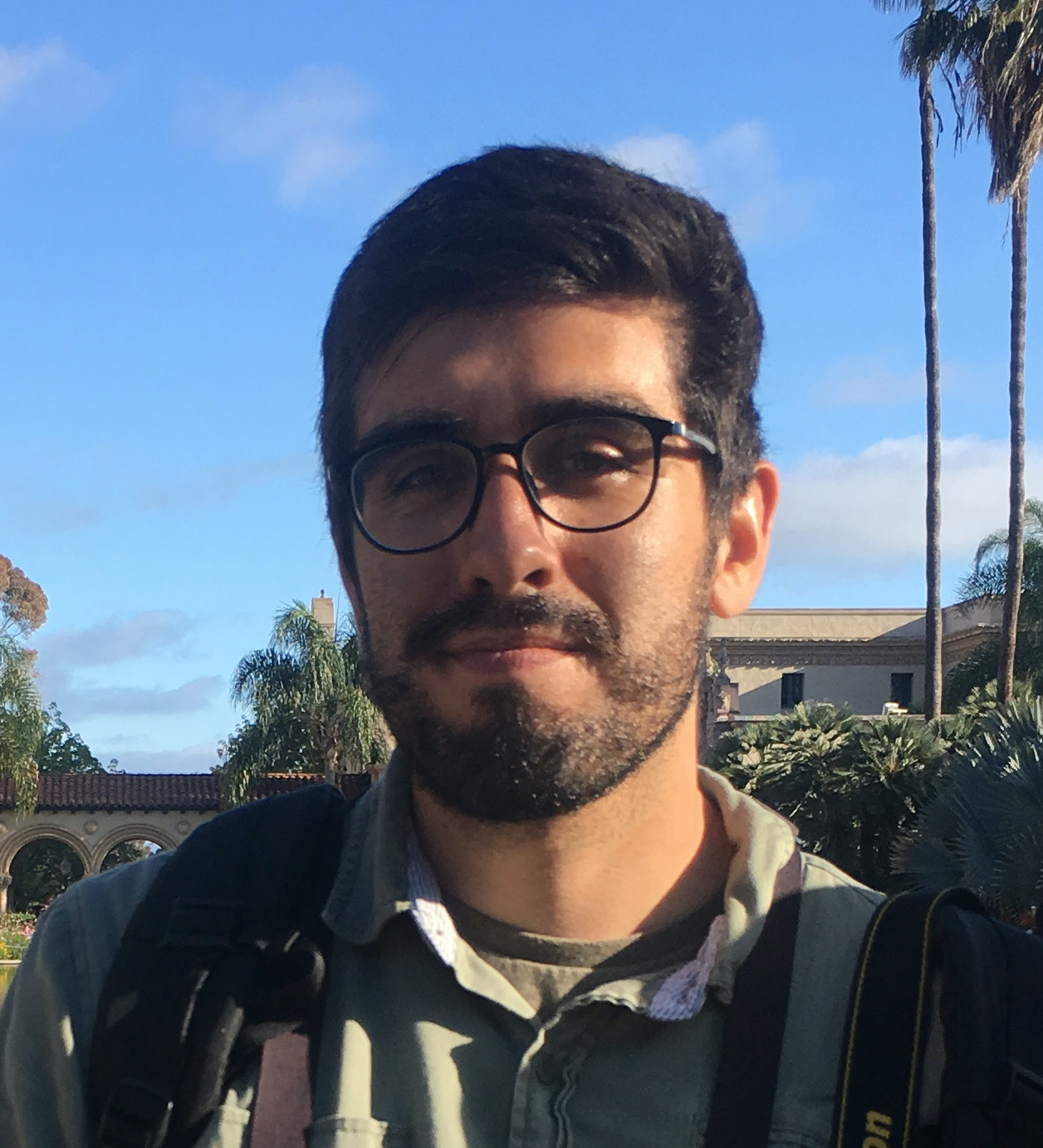}}]{Roberto Pineda} was born in Ensenada Baja California in 1991. He received the B.S. degree from CETYS Universidad, M\'{e}xico, in 2015, and the M.Sc. degree in computer science from CICESE, M\'{e}xico, in 2022. Roberto works as software engineer at ONEIL. His areas of interest include animal conservation through science, particularly bird conservation through salient object detection algorithms and the positive impact these could have on vulnerable bird species.
\end{IEEEbiography}

\begin{IEEEbiography}[{\includegraphics[width=1in,height=1.25in,clip,keepaspectratio]{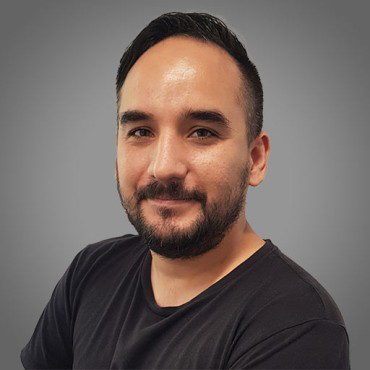}}]{Gerardo Ibarra} was born in Ciudad Victoria Tamaulipas in 1985. He received the B.S. degree in communications and electronics engineering from the Universidad Aut\'{o}noma de Tamaulipas, M\'{e}xico, in 2008, and the M.Sc. and Ph.D. degrees in electronics engineering from Universidad Aut\'{o}noma de San Luis Potos\'{\i}, M\'{e}xico, in 2010 and 2016, respectively. He is currently a postdoctoral researcher at the Institute for the Future of Education of Tecnol\'{o}gico de Monterrey, M\'{e}xico. His lines of research are robustness evaluation of image classification models, adversarial attacks in deep learning, deep learning in image processing, evolutionary algorithms for image classification, protruding object detection and interest point detectors.
\end{IEEEbiography}

\begin{IEEEbiography}[{\includegraphics[width=1in,height=1.25in,clip,keepaspectratio]{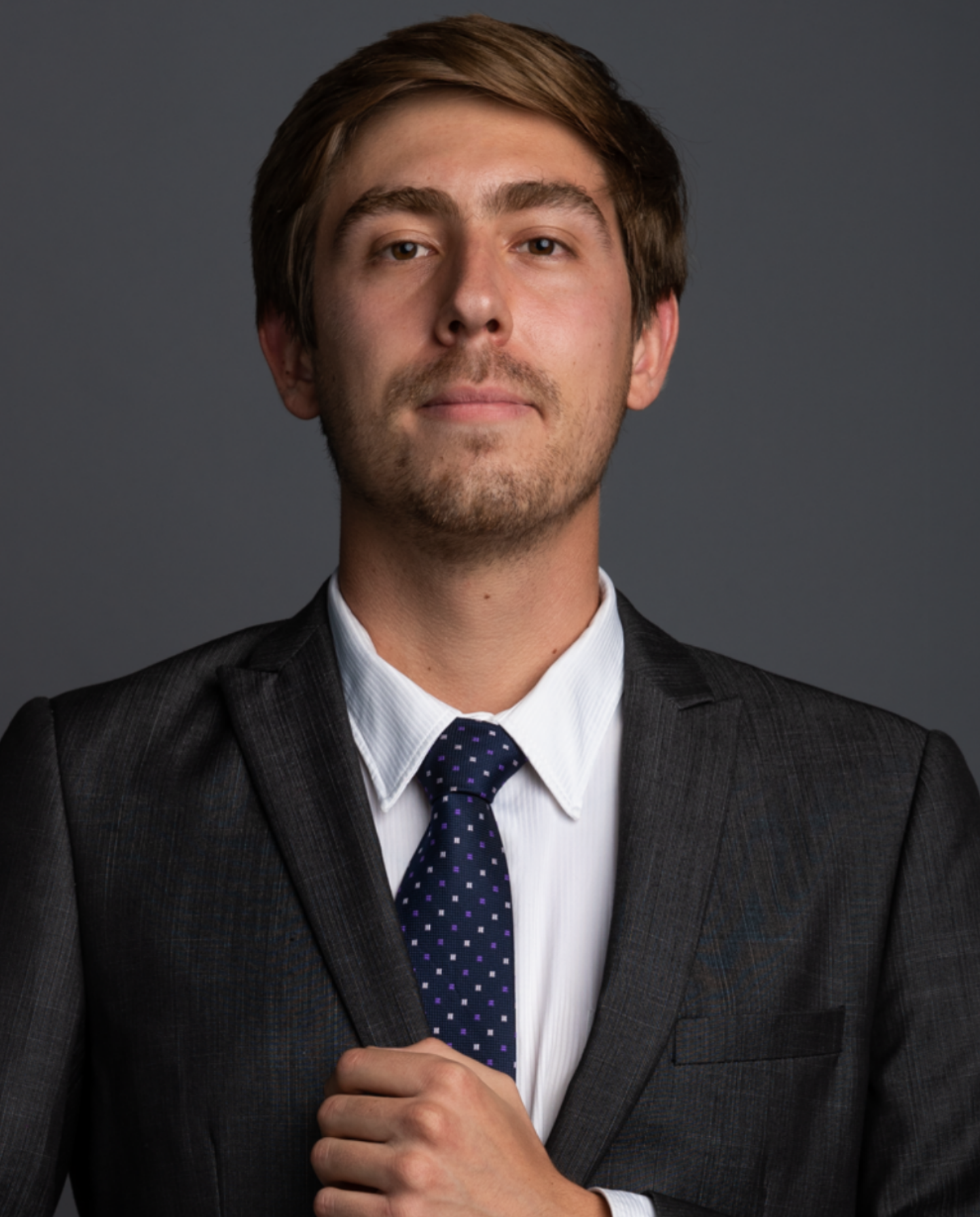}}]{Matthieu Olague} was born in Ensenada, Baja California, M\'{e}xico in 2000. He received the B.S. degree in Mechatronics from Universidad An\'{a}huac Qu\'{e}retaro, M\'{e}xico in 2022. Matthieu is a member of the directive board \& CTO of Intra. He leads a team of talented engineers constructing an ever-growing digital ecosystem, helping companies harness the power of digital. Before Intra, Matthieu developed software solutions and applications for various Mexican companies in collaboration with Data Pulse Analytics. In his spare time, Matthieu enjoys researching subjects related to Computer Vision and Machine Learning.
\end{IEEEbiography}

\begin{IEEEbiography}[{\includegraphics[width=1in,height=1.25in,clip,keepaspectratio]{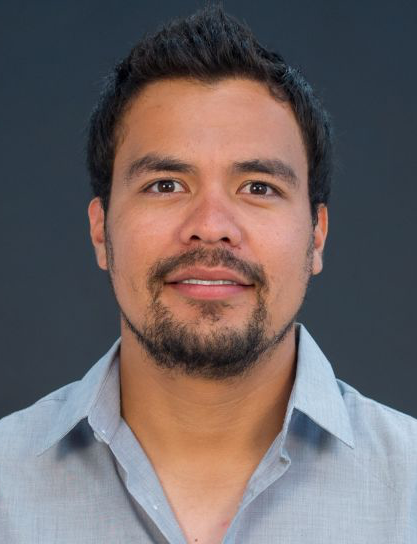}}]{Axel Martinez} was born in Distrito Federal, M\'{e}xico, in 1989. He received his B.S. degree in Mechatronics Engineering from Universidad An\'{a}huac M\'{e}xico Norte in 2017 and M.Sc. degree in Computer Science from CICESE in 2019. He is pursuing Ph.D. degree in Computer Science, and working at EvoVisi\'{o}n Laboratory at CICESE. His research interests are Computer Vision, Genetic Programming, Evolutionary Algorithms, 3D Reconstruction, and he enjoys surfing and cooking.
\end{IEEEbiography}

\begin{IEEEbiography}[{\includegraphics[width=1in,height=1.25in,clip,keepaspectratio]{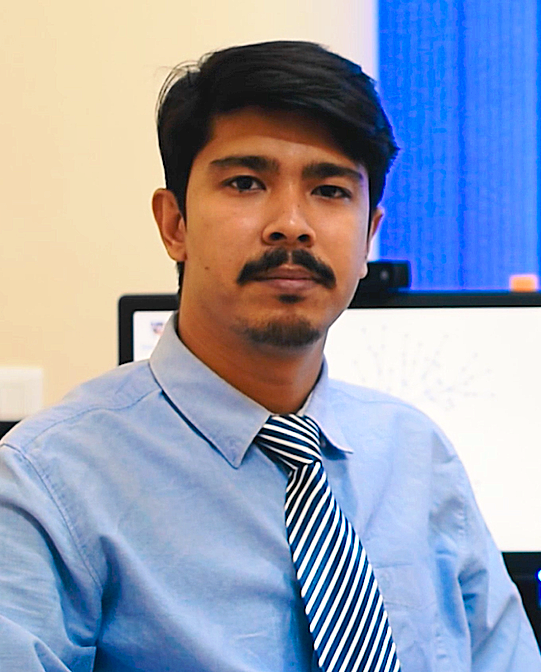}}]{Sambit Bakshi} is currently with the Department of Computer Science and Engineering, National Institute of Technology Rourkela, India. His area of interest includes surveillance, biometric security, and digital forensics. He served as Associate Editor of IEEE Access and Expert Systems in the past. He presently serves as associate editor of Innovations in Systems and Software Engineering Springer: A NASA Journal, Expert Systems with Applications, Image and Vision Computing, Multimedia Systems, and IEEE IT Professional magazine. He is a senior member of IEEE. He is Founding Chair of IEEE Rourkela Subsection. He presently serves as a member of IEEE Computational Intelligence Society Young Professionals Subcommittee since 2020 (liaison for IEEE Young Professional for the year 2022). He is a member of Professional Activities Committee of IEEE Region 10 for the year 2022. He has published widely in  more than 100 journals and conferences.  
\end{IEEEbiography}

\begin{IEEEbiography}[{\includegraphics[width=1in,height=1.25in,clip,keepaspectratio]{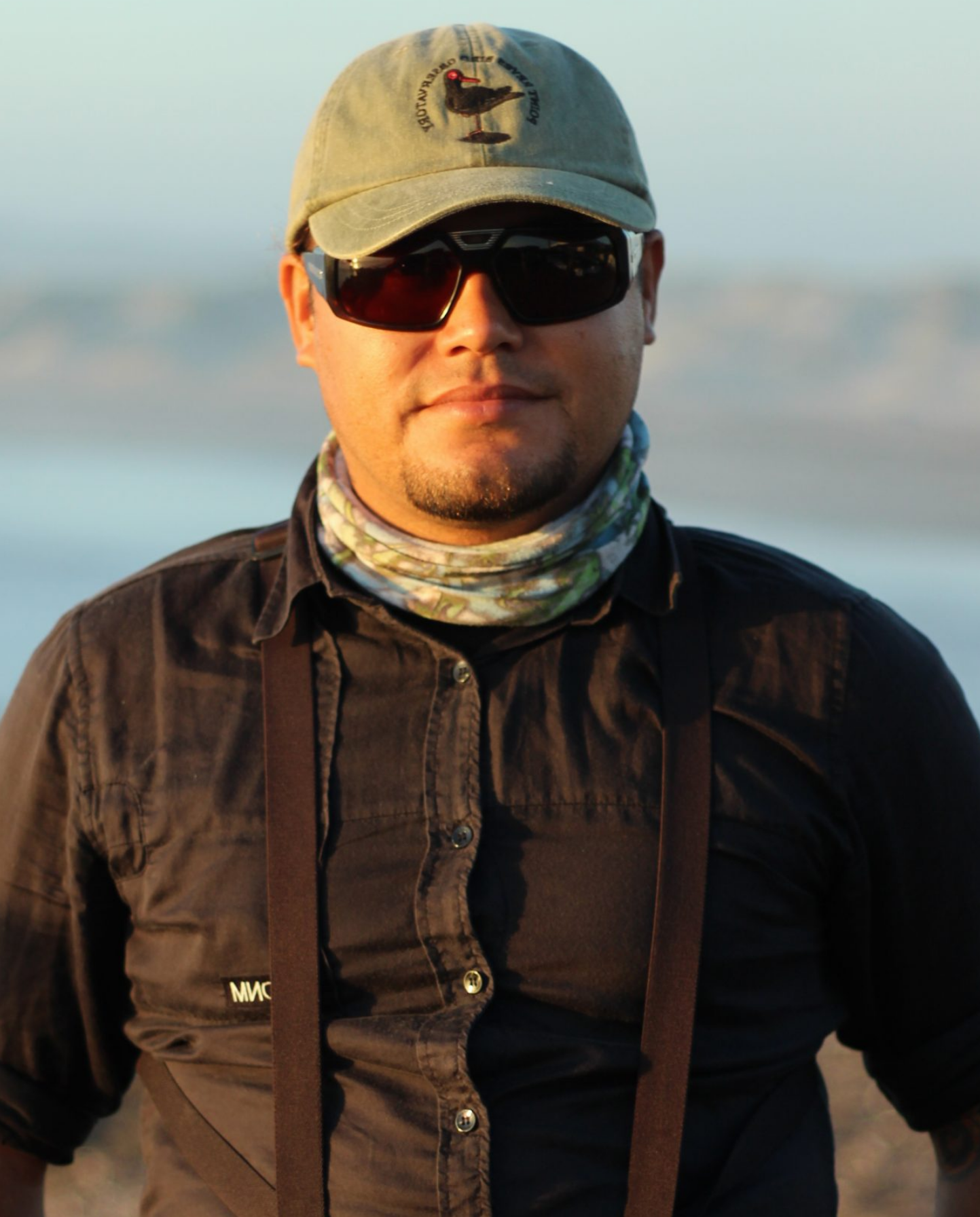}}]{Jonathan Vargas} was born in Nayarit in 1984. He received the B.S. degree in Biology from Universidad Aut\'{o}noma de Nayarit, M\'{e}xico, in 2012, and the M.Sc. degree in Marine and Coastal Sciences from Universidad Aut\'{o}noma de Baja California Sur, M\'{e}xico, in 2017. He is a Mexican biologist, conservationist and shorebird specialist and is based at Terra-Peninsular – an environmental NGO focused on conserving the biodiversity of the Baja California Peninsula. Jonathan has spearheaded the conservation of Todos Santos Bay, on the Pacific Coast of Baja California, where he started a shorebird-monitoring project that achieved the recognition of the bay as a site of regional importance for shorebirds and other migratory birds.
\end{IEEEbiography}

\begin{IEEEbiography}[{\includegraphics[width=1in,height=1.25in,clip,keepaspectratio]{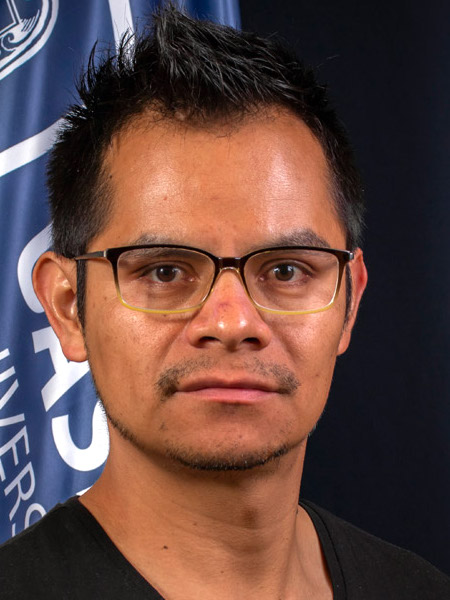}}]{Isnardo Reducindo} was born in M\'{e}xico City in 1985. He received the B.S. degree in communications and electronics engineering from the Universidad Aut\'{o}noma de Zacatecas, M\'{e}xico, in 2008, and the M.Sc. and Ph.D. degrees in electronics engineering from Universidad Aut\'{o}noma de San Luis Potos\'{\i}, M\'{e}xico, in 2010 and 2016, respectively. He is currently a Professor with the Information Science Faculty of Universidad Aut\'{o}noma de San Luis Potos\'{\i}, M\'{e}xico. His research interests include artificial intelligence in document management, digital humanities, technology in education, and information processing and management.  
\end{IEEEbiography}

\end{document}